\documentclass[11pt]{amsart}
\usepackage{geometry}                
\usepackage{graphicx,color}
\usepackage{amssymb}
\usepackage{epstopdf}
\usepackage{natbib}
\usepackage{multicol}

\usepackage[font=footnotesize]{caption}

\usepackage{amsthm}
\usepackage[foot]{amsaddr}

\newtheorem{thm}{Theorem}

\newcommand{\lp}{\left}
\newcommand{\rp}{\right}
\usepackage{hyperref}

\title{Optimistic Robust Optimization with Applications to 
  Machine Learning }
\author{Matthew Norton$^\dag$}
\address{$^\dag$Dept. of Operations Research, Naval Postgraduate School, Monterey, CA.}
\author{Akiko Takeda$^\ddag$$^\star$}
\address{$^\ddag$Dept. of Mathematical Analysis and Statistical Inference, The Institute of Statistical Mathematics, 
10-3 Midori-cho, Tachikawa, Tokyo 190-8562, Japan.}
\address{$^\star$RIKEN Center for Advanced Intelligence Project, 
1-4-1, Nihonbashi, Chuo-ku, Tokyo 103-0027, Japan}
\email{mnorton@nps.edu,atakeda@ism.ac.jp, sasha.mafusalov@gmail.com}
\author{Alexander Mafusalov}

\begin{document}

\maketitle
\begin{abstract}
Robust Optimization has traditionally taken a pessimistic, or worst-case viewpoint of uncertainty which is motivated by a desire to find sets of optimal policies that maintain feasibility under a variety of operating conditions. In this paper, we explore an optimistic, or best-case view of uncertainty and show that it can be a fruitful approach. We show that these techniques can be used to address a wide variety of problems. First, we apply our methods in the context of robust linear programming, providing a method for reducing conservatism in intuitive ways that encode economically realistic modeling assumptions. Second, we look at problems in machine learning and find that this approach is strongly connected to the existing literature. Specifically, we provide a new interpretation for popular sparsity inducing non-convex regularization schemes. Additionally, we show that successful approaches for dealing with outliers and noise can be interpreted as optimistic robust optimization problems. Although many of the problems resulting from our approach are non-convex, we find that DCA or DCA-like optimization approaches can be intuitive and efficient. 

\end{abstract}

\section{Introduction} 
Robust Optimization (RO) has proven to be one of the most popular methods for dealing with uncertainty in optimization. While taking many forms in various applications, see e.g. \cite{ben2009robust,bertsimas2011theory} for a thorough review, the traditional first-principal guiding RO is that of making pessimistic, worst-case assumptions regarding the realized outcomes of uncertain quantities. This first principal is well motivated, particularly given the primary goal of early RO techniques. Specifically, early RO techniques aimed to find sets of optimal policies that maintain feasibility (in the optimization sense) under a variety of operating conditions, see e.g. \cite{soyster1973technical,ben2000robust,bertsimas2004price}.

In this paper, we contend that the worst-case, pessimistic viewpoint is only half of the story. Specifically, we show that a best-case, optimistic viewpoint of uncertainty can be a fruitful approach to address a variety of problems. First, we find that this technique can be used in robust linear programming. Often times, robust linear programs that make only worst-case assumptions regarding uncertainty can prove to be overly conservative. For example, traditional RO assumes that the decision maker is powerless the uncertainty, with no resources available to combat or mitigate unfavorable realizations of uncertain parameters. We show that introducing optimistic, best-case uncertainty provides a method for reducing this conservatism in intuitive ways that encode economically realistic modeling assumptions, such as the availability of resources that can be used to combat or mitigate uncertainty. Furthermore, with this approach mixed-in with additional worst-case assumptions, we find that solutions are still robust in the traditional sense. After introducing the general approach in Section~\ref{sec_optrobstopt} and a general optimization strategy in Section~\ref{sec_solmethod}, we address robust linear programs in Section~\ref{sec_OR-LP}. Specifically, we demonstrate our methodology on the RO model of \cite{bertsimas2004price}, numerically demonstrating the efficacy of the approach.

Diverging from the linear programming context, we then show that the use of optimistic uncertainty is strongly connected to the machine learning literature. Pessimistic robust optimization has found many uses in machine learning applications, see e.g. \cite{el2003robust,bhattacharyya2005second,globerson2006nightmare,trafalis2007robust,xu2009risk,xu2009robustness,biggio2011support,katsumata2015robust}. However, most of these approaches take the traditional, worst-case view of uncertainty. For example, many convex regularization schemes have been shown to be equivalent to an assumption that data is subjected to worst-case uncertainty, e.g. see \cite{xu2009robustness}.

In a similar way, we show that popular techniques in machine learning can be interpreted as an application of an optimistic robust optimization approach with best-case uncertainty. We specifically address two contexts. First, we show that some popular non-convex regularizers can be interpreted through the lens of optimistic uncertainty. Sparse optimal solutions, e.g. sparse hyperplanes, can be extremely desirable in statistical decision making problems. For example, in classification and regression, sparsity often equates to variable selection while also helping to improve generalization. Recently, non-convex regularization has been shown to be a popular and effective technique for inducing sparsity in some applications. However, interpreting and understanding the exact effect that these regularizers have on optimal policies can be challenging. We show that inclusion of these regularizers is equivalent to assuming both best-case and worst-case data uncertainty, with optimism and pessimism working against each other in a very specific way. 

Beyond the issue of sparsity, we show that optimistic robust optimization provides a new view of methods in robust statistics.\footnote{In statistics, robustness usually refers to insensitivity to outliers, different from the concept of robust optimization assuming the data input to be uncertain and to belong to some fixed uncertainty set. There have been a large number of works on classical robust statistics, which develop estimation methods that are robust to outliers.} First, we address the situation where data is contaminated with outliers. We show that the effect of outliers can be intuitively addressed by making optimistic, best-case assumptions that discount the effect of the largest errors caused by outliers. We then show that our proposed formulation is strong connected, sometimes equivalent, to existing methods. Next, we consider the case where data might be contaminated by noise. It seems intuitive that one might benefit from an optimistic view of the data, attempting to see through the noise. We show that an existing method called the Total Support Vector Classifier (TSVC) from \cite{zhang2005support} is precisely equivalent to an optimistic robust approach. 

One of the major benefits of utilizing worst-case uncertainty assumptions is that it often yields convex, tractable reformulations. Thus, many RO problems are efficiently solvable and the introduction of worst-case uncertainty often does not add too much computational complexity to the nominal problem. In our paper, we find that optimistic, best-case uncertainty can be introduced similarly, forming tractable problems that, while non-convex, often have useful structure. Specifically, we consider a class of problems that when formulated with optimistic uncertainty can be reformulated as DC (Difference of Convex) optimization problems. Thus, we find that DC optimization methods like DCA (\cite{dinh2014recent}) can be quite effective and relatively efficient. 

The remainder of this paper is organized as follows. In Section~\ref{sec_optrobstopt}, we introduce the general formulation for optimistic robust optimization which contains both optimistic and pessimistic uncertainty. We then move to gain more specificity in Section~\ref{sec_TractableUS}, where we focus on problems that prove to have tractable reformulations as DC optimization problems. Section~\ref{sec_solmethod} overviews solution methods, focusing on DCA. In Section~\ref{sec_OR-LP}, we illustrate the use of our approach by modifying the popular RO model of \cite{bertsimas2004price}, while also presenting a numerical demonstration of its benefits. In Section~\ref{sec_applyML}, we utilize our approach in a machine learning context, discussing sparse non-convex regularization in Section~\ref{sec_sparseML} and prediction in the face of outliers and noisy data in Section~\ref{sec_outliersML}.

\section{Optimistic Robust Optimization} \label{sec_optrobstopt}
In this section, we introduce the general formulation of Optimistic Robust Optimization (ORO). First, we pose the problem with general functions and uncertainty sets. Then, we present the linear case, while also providing a simple trick for defining uncertainty sets that leads to tractable formulations. Of course, a wide variety of techniques can be used to define uncertainty sets (see \cite{ben2009robust,bertsimas2011theory} for a thorough treatment). Throughout this paper, however, we restrict the uncertainty sets in our examples to those induced by a norm. Specifically, letting $w \in \mathbb{R}^n$ denote our vector, we will make heavy use of $L_p$ norms, denoted by $\|w\|_p = \lp( \sum_i |w_i|^p \rp)^\frac{1}{p}$ for some $p \geq 1$. We will also frequently utilize a very flexible norm known as the \textit{largest-k} norm (also called the CVaR-norm from \cite{pavlikov2014cvar,mafusalov2013conditional} or D-norm from \cite{bertsimas2004robust}) which is given for $k \in [1,n]$ by
\[
\rho_k(w)=\sum_{i=1}^{\lfloor k \rfloor} |w^{(i)} | + (k - \lfloor k \rfloor)|w^{(\lceil k \rceil)}|,
\]
where $|w^{(1)}|\geq ...\geq |w^{(n)}|$ with superscripts denoting the ordered components of $w$ w.r.t. their absolute values, $\lceil k \rceil$ denoting the smallest integer that is greater than or equal to $k$, and $\lfloor k \rfloor$ denoting the largest integer that is less than or equal to $k$. This norm gives the sum of the largest $\lfloor k \rfloor$ components plus a part of the
$\lceil k \rceil$'th largest component when $k$ is not an integer.
The value $\rho_k(w)$ also can be obtained by solving an optimization problem:
\begin{equation}
   \label{cvar_opt}
  \rho_k(w)= \min_\zeta \zeta+\frac{1}{k}\sum_{i=1}^n [w_{i}-\zeta ]^+,
\end{equation}
where $[a]^+$ indicates $\max\{0,a\}$. It will also be useful to note that it has dual norm $\rho_k^*(w) = \max\{ \frac{\|w\|_1}{k} , \|w\|_\infty\}$

Additionally, we note that we approach uncertainty from the viewpoint of perturbations. Thus, instead of specifying explicitly an uncertainty set surrounding our uncertain parameter, we specify perturbations applied to our parameter, with these perturbations belonging to some specific set. Of course, these approaches are equivalent, but we find that the perturbation perspective is more intuitive for our purposes. 
\subsection{General Formulation} \label{sec_GeneralForm}
To introduce ORO, we begin with a general formulation. Specifically, let $w \in \mathbb{R}^n$ be our decision variable, let $X_i \in \mathbb{R}^m$ for $i=0,...,\ell$ be uncertain parameters or data, and let $f_i : \mathbb{R}^n \times \mathbb{R}^m \rightarrow \mathbb{R}$ be functions $i=0,...,\ell$. Assume, then, that we are given the \textit{nominal} problem where $X_i$ are fixed to some \textit{nominal} value\footnote{This could represent a single measurement or guess for the value of $X_i$.}:

\begin{equation}
\label{general_nominal}
\begin{aligned}
&\underset{w}{\text{min }}  
& & f_0 (w,X_0) \\
& s.t.  
& & f_i(w,X_i) \leq 0 , i=1,...,\ell \\
\end{aligned}
\end{equation}

Assume, however, that $X_i$ is subject to uncertainty which we represent as perturbations (or disturbances) $\delta_i \in C_i$, where $C_i \subset \mathbb{R}^m$ is some convex set. Typically, RO assumes that these disturbances are pessimistic, yielding the formulation:

\begin{equation}
\label{general_robust}
\begin{aligned}
&\underset{w}{\text{min }}   \underset{\delta}{\text{max }}  
& & f_0 (w,X_0+ \delta_0) \\
& s.t.  
& & f_i(w,X_i + \delta_i) \leq 0 , i=1,...,\ell \\
&&& \delta_i \in C_i , i=0,...,\ell
\end{aligned}
\end{equation}

Of course, in the context of traditional RO, the pessimistic viewpoint is well motivated. We want solutions $w$ that are feasible for a wide range of operating conditions, when $X_i$ may deviate from the nominal value. However, this can be overly pessimistic, focusing only on the worst case. In addition, the RO framework makes the critical assumption that we are powerless to fight back against uncertainty. Often times, however, the reality is much more optimistic. For example, in some contexts, we may have a budget or limited set of resources which we can use to combat uncertainty, effectively immunizing any issues which may arise when the true $X_i$ deviate from the nominal value. In other words, it may be well motivated to introduce some optimism about our ability to fight back against, deal with, or immunize \textit{some} of the uncertainty that may arise when we realize the true value of $X_i$. A more specific example and formulation which illustrates this motivation follows in Section~\ref{sec_OR-LP}.

For this reason, this work considers two types of uncertainty: \textit{pessimistic and optimistic}. Let $\delta^P$ denote the pessimistic disturbances and let $\delta^O$ denote the optimistic disturbances with respective convex sets $C^O, C^P \subset {R}^m$. We consider, generally speaking, formulations of the form:

\begin{equation}
\label{general_risky_robust}
\begin{aligned}
&\underset{w,\delta^O}{\text{min }}   \underset{\delta^P}{\text{max }}  
& & f_0 (w,X_0+ \delta^O_0 + \delta^P_0) \\
& s.t.  
& & f_i(w,X_i + \delta^O_i + \delta^P_i) \leq 0 , i=1,...,\ell \\
&&& \delta^O_i \in C^O_i , \delta^P_i \in C^P, i=0,...,\ell
\end{aligned}
\end{equation}

This formulation, and the introduction of optimistic uncertainty, allows us to encode less conservative views of uncertainty that are potentially well motivated by realistic assumptions regarding the ability of the decision maker to combat unfavorable realizations of the uncertain $X_i$. In addition, the way in which optimism is included in the formulation is optimal in the sense that one would obviously want to allocate the optimistic resources in a way that optimally counteracts pessimistic, worst-case outcomes so as to minimize losses (maximize profits).

As already mentioned, Section~\ref{sec_OR-LP} provides a more specific formulation to illustrate this point of intuition. However, as we will see in Section~\ref{sec_applyML}, this framework can often go beyond this basic intuition which is primarily motivated by economic contexts. Specifically, we see in machine learning applications that we can motivate the injection of optimism in a data-centric context, where we connect popular methods of non-convex regularization and robust statistics with optimistic robust optimization problems.

\subsection{Linear Programming Formulation and Tractable Uncertainty Sets} \label{sec_TractableUS}
To move our problem into a context that is easier to simplify into tractable formulations, we focus here on the specific context of Optimistic Robust Linear Programs (OR-LP) and a flexible type of uncertainty set which leads to tractable formulations. As the paper moves forward, we will simplify our uncertainty sets further to focus on norm-based uncertainty sets. 

Typically, when forming uncertainty sets for robust optimization problems, one starts by specifying an uncertainty set and then attempting to simplify the expression of uncertainty so that the optimization is tractable. For example, a natural notion of uncertainty is the norm based uncertainty. Then, using the idea of a dual norm, optimization over these sets reduces to optimization with the dual norm. We take a similar, but arguably more flexible approach that can also work in the other direction, where we can begin by specifying a convex function and formulate an uncertainty set based upon this function. The use of this other direction will become clear when we discuss regularization and applications.

Consider convex uncertainty sets $C(g,z)$ that are uniquely defined by a positive homogenous (PH)\footnote{A function $g:\mathbb{R}^n \rightarrow \mathbb{R}$ is positive homogenous if $g(0)=0$ and $ag(w)=g(aw)$ for any $a>0$.}, convex, non-negative function $g:\mathbb{R}^n \rightarrow \mathbb{R}$ and positive scalar $z \geq 0$. Specifically, since $g$ is convex, PH, and non-negative, we know from convex analysis (see e.g. Corollary 13.2.1 of \cite{rockafellar2015convex}) that $zg$ is the \textit{support function} for some convex set $C(g,z)$. We can think about this in two directions. We can begin with any convex set $C(g,z)$, and we can then solve for $zg$ (which we know exists since the set is convex) via the formula,
$$ zg(w)=\sup_{\delta \in C(g,z)} \delta^T w \;.$$
However, we can also work in the opposite direction. Given any convex, PH, and non-negative $zg$, we know that $zg$ is the support function for some convex set. Furthermore, we know that this convex set has a generic representation\footnote{See, again, Corollary 13.2.1 of \cite{rockafellar2015convex}.} given by,
$$C(g,z)=\{\delta | \delta^Tx \leq zg(x) \;, \forall \; x \in \mathbb{R}^n\} \;.$$

Moving onto the OR-LP formulation, we can demonstrate the use of defining uncertainty sets in this way. If we assume all $f_i$ from (\ref{general_risky_robust}) are linear and that the convex sets are given by $C(g_i^O,z_i^O), C(g_i^P,z_i^P)$, then our OR-LP is formulated as,

\begin{equation}
\label{LP_risky_robust1}
\begin{aligned}
&\underset{w,\delta^O}{\text{min }}   \underset{\delta^P}{\text{max }}  
& & w^T(X_0+ \delta^O_0 + \delta^P_0) \\
& s.t.  
& & w^T(X_i + \delta^O_i + \delta^P_i) \leq 0 , i=1,...,\ell \\
&&& \delta^O_i \in C(g_i^O,z_i^O) , \delta^P_i \in C(g_i^P,z_i^P), i=0,...,\ell
\end{aligned}
\end{equation}

The discussion above on the support function of $C(g,z)$ leads to the following theorem.
\begin{thm} \label{theo_optrobust_nonconv1}
  When $C(g,z)$ is defined with a PH, convex, non-negative $g$ and $z \geq 0$, we can simplify the formulation \eqref{LP_risky_robust1} into the equivalent,
\begin{equation}
\label{LP_risky_robust2}
\begin{aligned}
&\underset{w}{\text{min }}  
& & w^TX_0+ z_0^P g_0^P(w) - z_0^O g_0^O(w) \\
& s.t.  
& & w^TX_i + z_i^P g_i^P(w) - z_i^O g_i^O(w) \leq 0 , i=1,...,\ell 
\end{aligned}
\end{equation}
\end{thm}

Theorem~\ref{theo_optrobust_nonconv1}, and most of the results we discuss in this paper, rely upon the positive homogeneity of $g$. However, we point out in Theorem~\ref{theo_optrobust_nonconv2} that when \eqref{LP_risky_robust2} is formulated with function $g$ that is not PH, but is still closed and convex, it is possible to formulate an equivalent robust formulation that is similar to \eqref{LP_risky_robust1}. We relegate this result to the appendix since we do not make heavy use of it throughout our discussion.

The formulations already presented are meant to provide a general feel for our approach. In the following sections, we will motivate the use of optimistic uncertainty in the linear programming context followed by a machine learning context. First, in Section~\ref{sec_OR-LP} we show how this approach can be applied to the popular formulation of \cite{bertsimas2004price} for robust linear programming. This allows us to pose optimistic uncertainty in terms that have real economic significance. Second, in Section~\ref{sec_applyML} we discuss applications of this framework in machine learning. We show that non-convex regularization can often be presented within this framework, providing new geometric intuition to explain its success in inducing sparsity. We also briefly discuss the use of optimistic uncertainty for combating problematic data sets, such as those containing outliers, making connections to methods of robust statistics. Throughout our experiments, we utilize DCA to solve our resulting formulations. Section~\ref{sec_solmethod} presents a brief overview of this procedure and comments upon its effectiveness in our considered examples. Overall, we find that even the simplest implementation of DCA is both heuristically appealing and can achieve quality solutions. We then point to work which performs more efficient, advanced variants of DCA for specific problems. It should be noted that readers can, if desired, skip ahead to the application discussion in Section~\ref{sec_OR-LP}-\ref{sec_applyML}, leaving Section~\ref{sec_solmethod} for last without significant issue.

\section{Solution Methods} \label{sec_solmethod}
In this section, we overview efficient methods for solving the non-convex problems of ORO.  Although these methods often include solving a convex subproblem within each iteration, reasonable solutions can often be found after very few iterations. Section~\ref{sec_OR-LP} briefly illustrates this point within a numerical demonstration in a linear programming setting.

In this paper, we have focused our attention on ORO problems that often reduce to DC optimization problems. We begin by discussing OR-LP's. The solution method we apply, called DCA, is representative of the general methodology we apply to other ORO problems. For the unconstrained problems mentioned in Section~\ref{sec_applyML}, although most do not have an explicit DC decomposition that we are yet aware of, we know that the function is, in fact, DC. Thus, we apply the methodology of DCA, albeit not exactly. We discuss this general approach after discussing OR-LP's.
\subsection{Solving OR-LP's} \label{sec_algo_OR-LP}
To solve any general OR-LP formulated as (\ref{LP_risky_robust1}), we can utilize DC optimization techniques due to the fact that (\ref{LP_risky_robust1}) reduces to the DC program (\ref{LP_risky_robust2}). In general, it is useful to pose the DC algorithm as an alternating optimization approach to the following problem where we alternate between optimizing $w$ and $\delta^O$:
\begin{equation}
\label{LP_risky_robust3}
\begin{aligned}
&\underset{w,\delta^O}{\text{min }}   
& & w^TX_0+ z_0^Pg_0^P(w)  + w^T \delta^O_0 \\
& s.t.  
& & w^TX_i +z_i^Pg_i^P(w)  + w^T \delta^O_i  \leq 0 , i=1,...,k \\
&&& \delta^O_i \in C(g_i^O,z_i^O) ,  i=0,...,k
\end{aligned}
\end{equation}

Following the general approach outlined in \cite{dinh2014recent}, we utilize the following algorithm:

\begin{enumerate}
\item Initialize: Number of iterations or convergence criteria.
\item Initialize: $w$ equal to argmin of (\ref{LP_risky_robust3}) with fixed $\delta_i^O=0$ for every $i=0,...,k$.
\item Set $\hat{\delta}_i^O  \in \underset{  \delta \in C(g_i^O,z_i^O) }{ \text{argsup }} w^T \delta$ for all $i=0,...,k$.
\item Set $w$ equal to argmin of (\ref{LP_risky_robust3}) with fixed $\delta_i^O=-\hat{\delta}_i^O$ for every $i=0,...,k$.
\item If not converged or iteration limit not reached, return to Step 3.
\end{enumerate}

It is easy to see that this algorithm is equivalent to DCA by recognizing that, if $\partial \left( zg(w) \right)$ denotes the subdifferential of $zg$ at $w$, then
 $$\delta \in \partial \left(zg(w) \right) \iff \delta  \in \underset{  \delta \in C(g,z) }{ \text{argsup }} w^T \delta  \iff zg(w) =\underset{  \delta \in C(g,z) }{ \text{sup }} w^T \delta \;.$$
This fact follows easily from the definition of a support function. Therefore, we see that Step 3 is solving for a subgradient and Step 4 is solving the original problem, but with the concave part linearized via a subgradient. In addition, this perspective shows that it is not necessary to solve the optimization problem in Step 3 if a subderivative of $g$ can be found via more efficient means, e.g. a closed form solution. This approach is also heuristically appealing, meaning that if one had no knowledge of the convergence guarantees of DCA, the proposed optimization method could be justified as a type of greedy, step-wise procedure. The initial convex problem in Step 2 ignores the optimistic uncertainty, solving a robust LP. As the algorithm progresses, we are then injecting our optimistic beliefs to relax the conservatism of the initial robust problem.

\subsection{Solving General ORO Formulations}
In general, DCA is applied to DC problems which have an exact DC decomposition. For OR-LP's, this decomposition was easy to come by because of the separability of the linear function, meaning that $w^T(X+\delta^P+\delta^O)=w^TX + w^T\delta^P + w^T\delta^O$. However, in the general case of (\ref{general_risky_robust}), the function $f(w,X+\delta^P+\delta^O)$ may be DC but not have an obvious DC decomposition. Nevertheless, we show here that we can still apply the DCA methodology in an appealing way to certain DC problems that may not have a known DC decomposition.

Considering formulation (\ref{general_risky_robust}), let us assume that for every $i=1,...,k$, we have that $f_i(w,X)=h_i(w^TX)$ for some convex and non-decreasing $h_i$. We then would have that
\begin{align*}
 \underset{\delta^O \in C(g^O,z^O) }{\text{inf}}   \underset{\delta^P \in C(g^P,z^P)  }{\text{sup}} f_i(w,X+\delta^P+\delta^O) &=  \underset{\delta^O \in C(g^O,z^O) }{\text{inf}}   \underset{\delta^P \in C(g^P,z^P)  }{\text{sup}} h_i(w^T(X+\delta^P+\delta^O) ) \\
&=  \underset{\delta^O \in C(g^O,z^O) }{\text{inf}}   \underset{\delta^P \in C(g^P,z^P)  }{\text{sup}} h_i( w^TX + w^T \delta^P + w^T \delta^O) \\
&= h_i( w^TX + z^P g^P(w) - z^Og^O(w) )
 \;,
 \end{align*}
where the last equality holds since $h_i$ is non-decreasing. Since $h_i, g^O, g^P$ are convex, under mild conditions\footnote{See e.g. the discussion regarding composition of DC functions in Section 2 of \cite{hartman1959functions} or Chapter 4 of \cite{tuy1998convex}. },we will have that $h_i( w^TX + z^P g^P(w) - z^Og^O(w) )$ is DC.
Therefore, we can reformulate our problem as (\ref{RRO_sep}) and apply the same alternating minimization algorithm as was done for OR-LP's, but applied to (\ref{RRO_sep}) instead of (\ref{LP_risky_robust3}).

\begin{equation}
\label{RRO_sep}
\begin{aligned}
&\underset{w,\delta^O}{\text{min }}   
& & h_0(w^TX_0+ z_0^Pg_0^P(w)  + w^T \delta^O_0) \\
& s.t.  
& & h_i(w^TX_i +z_i^Pg_i^P(w)  + w^T \delta^O_i ) \leq 0 , i=1,...,k \\
&&& \delta^O_i \in C(g_i^O,z_i^O) ,  i=0,...,k
\end{aligned}
\end{equation}

Thus, we are essentially following the same procedure as before, where we are linearizing the concave part via subgradients and solving the resulting convex subproblems. We will see that many of the formulations presented in further sections are special cases of (\ref{RRO_sep}). Therefore, we apply this DCA-like algorithm to perform the optimization. 
 
\subsection{Faster Methods}
In this paper, because of the generality of our formulation, we utilize the very simple DCA procedure which requires solving a convex subproblem at every iteration. However, it is important to note that some DC optimization problems can be decomposed in a way such that the convex subproblem of DCA has a closed form solution. This allows DCA to avoid the bottleneck of having to repeatedly apply a convex optimization procedure within each iteration. A example which is related to our discussion in Section~\ref{sec_sparseML} is the Proximal-DCA procedure utilized by \cite{tono2017efficient} and \cite{gotoh2017dc}. In short, these papers show that a specific class of estimation problems, when formulated with a particular sparsity inducing non-convex regularizer, can be decomposed so the problem is DC and, importantly, so that the solution of the convex subproblem of DCA is equivalent to a proximal operator which has a closed form solution. Thus, even though we find the simple DCA procedure sufficient for our purposes in general, there are proven methods for improving the efficiency of the optimization procedure for specific applications and formulations.

\section{Optimistic Robust Linear Programming With Budgets of Uncertainty} \label{sec_OR-LP}

In this section we illustrate application of optimistic robust optimization, primarily focusing our analysis on Optimistic Robust Linear Programs (OR-LP's) having a budget-restricted uncertainty while also attempting to illustrate how one can intuitively introduce optimistic uncertainty into traditional RO frameworks.
The concept of a {\it budget of uncertainty} was introduced by  \cite{bertsimas2004price} for constructing a type of polyhedral uncertainty set. This uncertainty set is defined by the largest-$k$ norm and the parameter $k$ indicates a level of flexibility in choosing the tradeoff between robustness and performance. We will introduce one more parameter $r$ as the optimistic parameter to mitigate excessive robustness, which leads to a non-convex optimization formulation.

\subsection{Ordinary Robust Linear Programming}
We begin by considering the general robust LP discussed in \cite{bertsimas2004price}. In their problem, they begin by assuming that there is some nominal LP given by:

\begin{equation}
\label{LP_nominal}
\begin{aligned}
&\underset{w}{\text{min }}  
& & w^T c \\
& s.t.  
& & w^T a_j \leq b_j , \forall j=1,...,\mathcal{J} \\
& & & l \leq w \leq u
\end{aligned}
\end{equation}
In the following, we assume the uncertainty for the coefficients $b_j, l , u,$ and $c$ to be deterministic. However, it does not limit the generality of the problem in the sense that uncertainty in $c$ can be reflected by the addition of the constraint $w^T c \leq t$ with the objective replaced by deterministic $t$ and uncertainty in $b_j,l,$ or $u$ can be treated in the same manner as uncertainty in $a_j$ is treated. 

\cite{bertsimas2004price} assume that the given coefficients $a_{ij}$ are only \textit{nominal values} and that their true value $\hat{a}_{ij} $ is uncertain, specifically a symmetric random variable distributed on the interval $[ a_{ij} - \bar{a}_{ij} , a_{ij} + \bar{a}_{ij}]$ where $\bar{a}_{ij}$ is some known maximal variation from the nominal value. They then take a pessimistic approach to find a policy $w$ that is feasible with high probability for realizations of $\hat{a}_{ij}$ by solving the following pessimistic, robust LP with free parameters $k_j \in [1,n]$ and diagonal matrix $\bar{A_j}=diag\{\frac{1}{\bar{a}_{1j}},...,\frac{1}{\bar{a}_{nj}} \} $:

\begin{equation}
\label{LP_robust1}
\begin{aligned}
&\underset{w}{\text{min }}  
& & w^T c \\
& s.t.  
& &  \sup_{\delta_j^P} w^T (a_j + \delta_j^P) \leq b_j , \forall j=1,...,\mathcal{J} \\
& & & \| \bar{A_j} \delta_j^P \|_1 \leq k_j  \\
& & & \| \bar{A_j} \delta_j^P \|_\infty \leq 1  \\
& & & l \leq w \leq u \;.
\end{aligned}
\end{equation}

This formulation is somewhat difficult to understand. However, we can simplify to (\ref{LP_robust2}) by recognizing that the uncertainty set is related to the norm $\rho_k^*(w) = \max\{ \frac{\|w\|_1}{k} , \|w\|_\infty\}$ which is the dual of the \textit{largest-k} norm $\rho_k(w) = \sum_{i=1}^{\lfloor k \rfloor} |w^{(i)} | + (k - \lfloor k \rfloor)|w^{(\lceil k \rceil)}|
$.

\begin{equation}
\label{LP_robust2}
\begin{aligned}
&\underset{w}{\text{min }}  
& & w^T c \\
& s.t.  
& &  w^T a_j + \rho_{k_j}( \bar{A}_j^{-1} w ) \leq b_j , \forall j=1,...,\mathcal{J} \\
& & & l \leq w \leq u \;
\end{aligned}
\end{equation}
Clarifying further, we can also write the robust constraint as a sum. First, let $y_{ij}=\bar{a}_{ij}w_i$ so that $y_j=\bar{A}_j^{-1} w$. Next, denote the ordered components (w.r.t. absolute value) of $y_j$ as $|y_j^{(1)}| \geq \dots \geq |y_j^{(n)}|$. Then, finally, the robust constraint can be written as
$$w^T a_j +\sum_{i=1}^{\lfloor k_j \rfloor} |y_j^{(i)} | + (k_j - \lfloor k_j \rfloor)|y_j^{(\lceil k_j \rceil)}|  \leq b_j \;.$$
Put simply, for a given $w$, this formulation protects against the worst-case $k_j$ components of $\hat{a}_j$. Thus, if less than $k_j$ components of $\hat{a}_j$ deviate from their nominal value $a_{ij}$, the optimal solution $w$ obtained via (\ref{LP_robust2}) will not violate the $j^{th}$ constraint.

Much of the motivation and discussion in \cite{bertsimas2004price} is geared toward achieving solutions that are less conservative than the robust policies from \cite{ben2000robust} and \cite{soyster1973technical}, while still maintaining a high degree of probable feasibility under many conditions (this is their motivation for introducing the free parameter $k_j$). In some sense, the entire goal is two-fold. First, to maintain a high degree of robustness. Second, to obtain a smaller objective (in the minimizing sense) by being less conservative. Here, we show that optimistic uncertainty can easily be used to achieve this effect. We also do this in an intuitive way that can be directly motivated by apriori knowledge of the actual problem. 

\subsection{Price of Optimistic Robustness}
The formulation from \cite{bertsimas2004price} protects against the worst case $k_j$ components of $\hat{a}_j$ relative to a given solution $w$. Assume, though, that we have resources so that for any plan (or decision vector) $w$, we can alter $r_j<k_j$ components of $\hat{a}_j$, for every $j$, so that these components are guaranteed to remain at their nominal values $a_{ij}$ (i.e. we have resources that can counteract or prevent harmful deviations from the nominal value for $r_j$ components of coefficient vector $\hat{a}_j$).\footnote{The more realistic interpretation is that we can use our optimistic resources to counteract deviations from the nominal value (or the affects of such a deviation) \textit{after} they have been realized. Of course, for our formulation, this is equivalent to assuming that we can guarantee no deviation from the nominal by applying our resources, even though this interpretation is less convincing.}  Thus, we assume that we can be optimistic about the $r_j$ worst components of each vector $\hat{a}_j$. This, essentially, amounts to an assumption that resources are available to counteract a limited set of unexpected circumstances. For example, if coefficients were to represent resource requirements for some product, then investment in new equipment and technology, or a change in product requirements can be used to protect against harmful deviations from the nominal value $a_{ij}$. Thus, the optimistic assumption is completely realistic and can be directly understood in terms of the real economic resources available to the decision maker. Furthermore, if these resources were available, we would certainly want to know what the optimal allocation strategy would be. We would want to invest in eliminating harmful coefficient deviations of the $r_j$ coefficients that lead to maximal profit (or minimal losses). Additionally, we will still have a level of robustness because we are assuming the worst-case for $k_j-r_j$ components of $\hat{a}_j$, i.e. since resources were limited, we were not able to counteract a set of $k_j-r_j$ component deviations from the nominal value.

Putting these assumptions into the formulation, we have the following OR-LP, where we assume that $r_j<k_j$.
\begin{equation}
\label{RRLP_1}
\begin{aligned}
&\underset{w}{\text{min }}  
& & w^T c \\
& s.t.  
& &  \inf_{\delta_j^O} \sup_{\delta_j^P} w^T (a_j + \delta_j^P + \delta_j^O) \leq b_j , \forall j=1,...,\mathcal{J} \\
& & & \| \bar{A_j} \delta_j^P \|_1 \leq k_j  \\
& & & \| \bar{A_j} \delta_j^P \|_\infty \leq 1  \\
& & & \| \bar{A_j} \delta_j^O \|_1 \leq r_j  \\
& & & \| \bar{A_j} \delta_j^O \|_\infty \leq 1  \\
& & & l \leq w \leq u \;.
\end{aligned}
\end{equation}
This problem then reduces to:
\begin{equation}
\label{RRLP_2}
\begin{aligned}
&\underset{w}{\text{min }}  
& & w^T c \\
& s.t.  
& &  w^T a_j + \rho_{k_j}( \bar{A_j}^{-1} w )  -  \rho_{r_j}(  \bar{A_j}^{-1} w )  \leq b_j , \forall j=1,...,\mathcal{J} \\
& & & l \leq w \leq u \;.
\end{aligned}
\end{equation}
The meaning of this formulation becomes clearer when considering the following equality, where $y_j$ is defined as before:
\begin{align*}
 \rho_{k_j}( \bar{A_j}^{-1} w )  -  \rho_{r_j}( \bar{A_j}^{-1} w ) 
 & = \sum_{i=1}^{\lfloor k_j \rfloor} |y_j^{(i)} | + (k_j - \lfloor k_j \rfloor)|y_j^{(\lceil k_j \rceil)}|    \\
 &- \left( \sum_{i=1}^{\lfloor r_j \rfloor} |y_j^{(i)} | + (r_j - \lfloor r \rfloor)|y_j^{(\lceil r_j \rceil)}| \right)   \\
 & = \sum_{i=\lfloor r_j \rfloor+1}^{\lfloor k_j \rfloor} |y_j^{(i)} | + (k_j - \lfloor k_j \rfloor)|y_j^{(\lceil k_j \rceil)}|   - (r_j - \lfloor r_j \rfloor)|y_j^{(\lceil r_j \rceil)}|  \;.\\
 \end{align*}
Thus, for each realized constraint vector $\hat{a}_j$, we assume that we can immunize $r_j$ of the worst case coefficients $\hat{a}_{ij}$, $i=1,...,n$, and we further assume the worst case for the other $k_j-r_j$ coefficients to keep us robust.

\subsection{Numerical Demonstration}
To demonstrate numerically the behavior of solutions for (\ref{RRLP_2}) as $r_j$ changes, we analyze solutions for two data sets. First, we use the real CAPRI data set from the NETLIB library of LP test problems which has dimension $w \in \mathbb{R}^{353}$, with $\mathcal{J}=129$ inequality constraints, and $142$ equality constraints.\footnote{CAPRI also imposes upper and lower bounds on decision variables $w_i$ which we did not subject to uncertainty.} Second, we generate a random LP with dimension $w \in \mathbb{R}^{250}$, zero equality constraints, and $\mathcal{J}=50$ inequality constraints.\footnote{Additionally, we enforced $w \geq 0$ and did not impose uncertainty on this constraint.}

For the CAPRI data set, we assumed that only the coefficients within each inequality constraint were subject to variation within $2\%$ of their nominal values. Then, for each row $j$ that was subject to uncertainty, we set $k_j$ as a fixed percentage of the non-zero coefficients within that row. Similarly, we set $r_j$ as a fixed percentage of $k_j$. Thus, let $\beta_j$ denote the number of non-zero coefficients in row $j$ and let $K \in \{0,.05,...,.95,1\}$ and $R \in \{0,.05,...,.25\}$ denote our percentages. We then say that we solved (\ref{RRLP_2}) for a \textit{pair} $(K,R)$ when we set $k_j= K\beta_j$ and $r_j =Rk_j$ for every row $j$. Thus, we solved (\ref{RRLP_2}) a total of $21\times 6$ times, once for each pair $(K,R)$, with $K \in \{0,.05,...,.95,1\}$ and $R \in \{0,.05,...,.25\}$. 

For the synthetic data set, we sampled the coefficient matrix $A\in \mathbb{R}^{50 \times 250}$ from a uniform $U(0,.5)$ distribution and cost coefficients $c$ from a uniform $U(0,1)$ distribution, setting $b=\mathbf{1}$. We include a non-negativity constraint $w \geq \mathbf{0}$ with no equality constraints or upper bounds on $w$. Furthermore, we assumed that coefficients $A$ were subject to variation within $10\%$ of their nominal values. Similar to the CAPRI setup, we then solved (\ref{RRLP_2}) using the algorithm shown in Section~\ref{sec_algo_OR-LP}
a total of 150 times, once for each pair $(K,R)$, with $K \in \{0,.01,...,.24\}$ and $R \in \{0,.05,...,.25\}$ denoting our percentages. 

\begin{figure}
\centering
\begin{minipage}{.5\textwidth}
  \centering
  \captionsetup{width=.9\linewidth}
  \includegraphics[width=3.24in,height=2.5in]{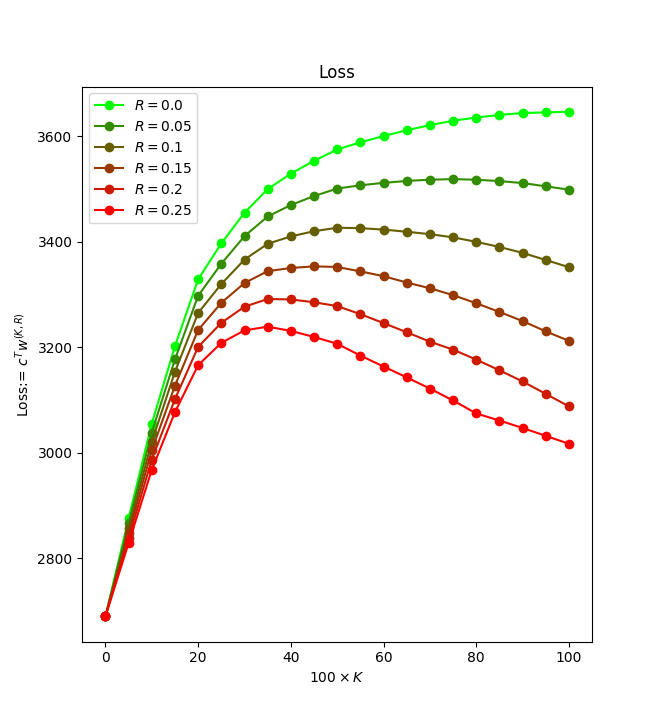} 
 \captionof{figure}{CAPRI: Loss (Optimal value) vs. $100 \times K$ [\%]}
  \label{fig_1}

\end{minipage}%
\begin{minipage}{.5\textwidth}

  \centering
  \captionsetup{width=.9\linewidth}
  \includegraphics[width=3.25in,height=2.5in]{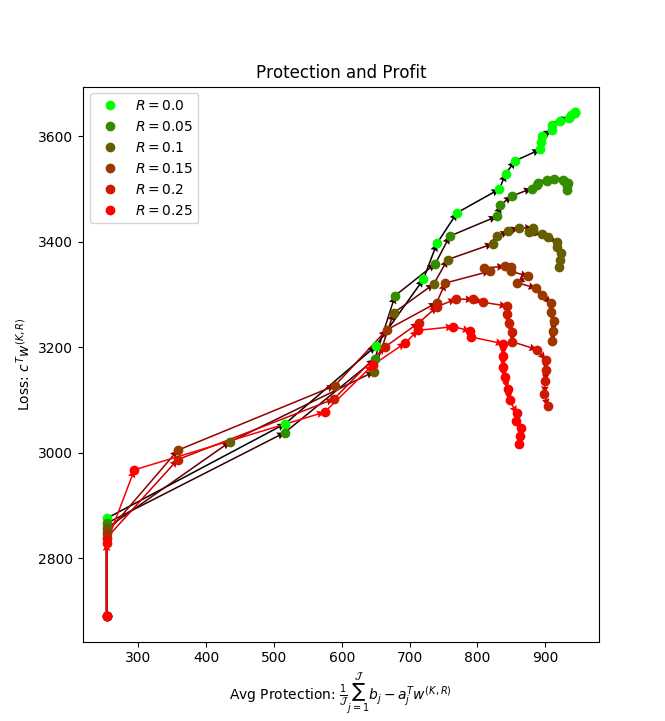} 
  \captionof{figure}{CAPRI: Loss (Optimal value) vs.  Avg protection over all constraints for all solutions $w^{(K,R)}$
    ($\frac{1}{\mathcal{J}}\sum (b_j - a_j^T w^{(K,R)})$). Arrows indicate increasing $K$ for fixed $R$.}
    \label{fig_2}

\end{minipage}

\end{figure}

\begin{figure}
\centering
\begin{minipage}{.5\textwidth}
  \centering
  \captionsetup{width=.9\linewidth}
  \includegraphics[width=3.24in,height=2.5in]{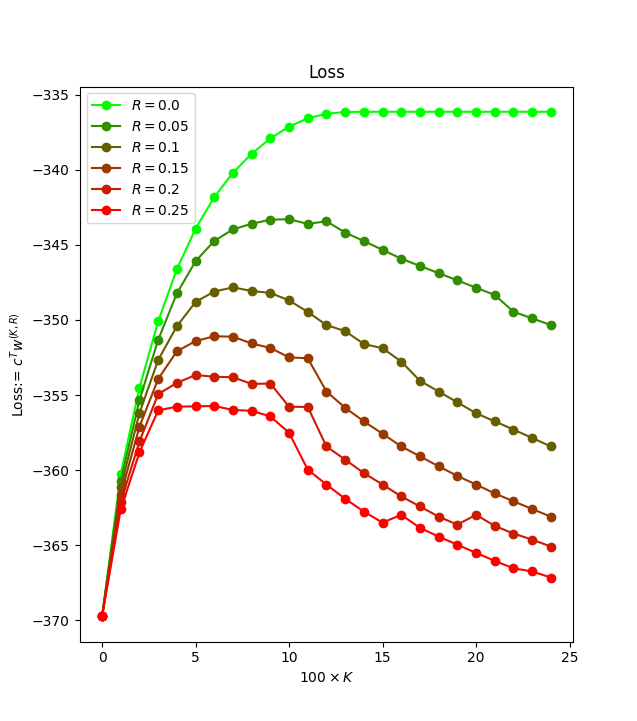} 
 \captionof{figure}{RANDOM: Loss (Optimal value) vs. $K$}
  \label{fig_A1}

\end{minipage}%
\begin{minipage}{.5\textwidth}

  \centering
  \captionsetup{width=.9\linewidth}
  \includegraphics[width=3.25in,height=2.5in]{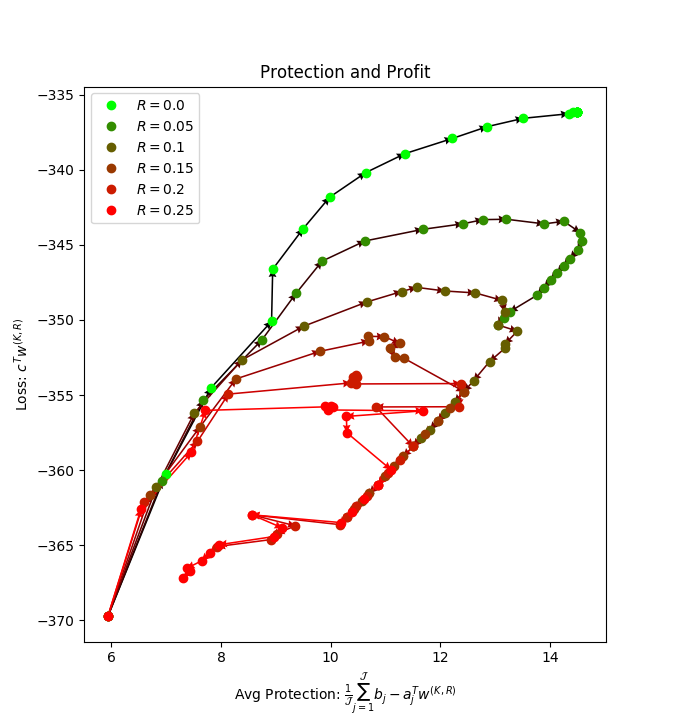} 
  \captionof{figure}{RANDOM: Loss (Optimal value) vs. Avg protection over all constraints for all solutions $w^{(K,R)}$ ($\frac{1}{\mathcal{J}}\sum (b_j - a_j^T w^{(K,R)})$). Arrows indicate increasing $K$ for fixed $R$.}
    \label{fig_A2}

\end{minipage}

\end{figure}

 \begin{figure}
  \centering
  \captionsetup{width=.9\linewidth}
  \includegraphics[width=6in,height=4in]{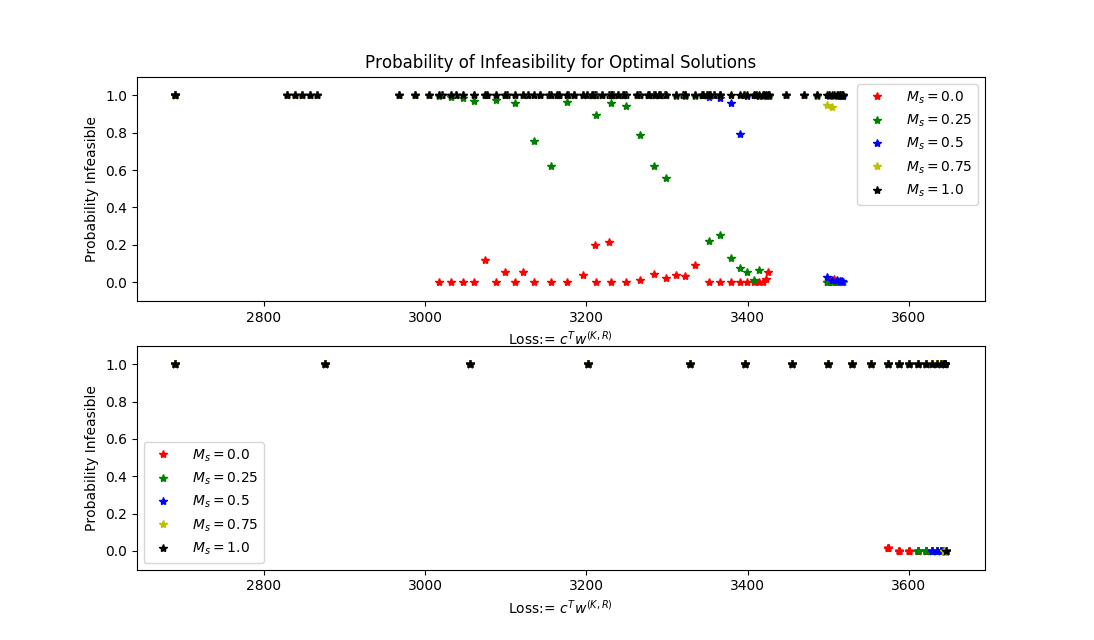} 
 \captionof{figure}{CAPRI: Upper chart plots loss vs. estimated $P(\hat{A}w^{(K,R)} > b)$ for all solutions $w^{(K,R)}$ with $R\neq 0$ for all scenarios $s\in \{1,...,5\}$. Lower chart plots the same for all solutions $w^{(K,R)}$ with $R= 0$. $M_s$ indicates scenario multiplier.}
   \label{fig_3}

\end{figure}

 \begin{figure}
  \centering
  \captionsetup{width=.9\linewidth}
  \includegraphics[width=6in,height=4in]{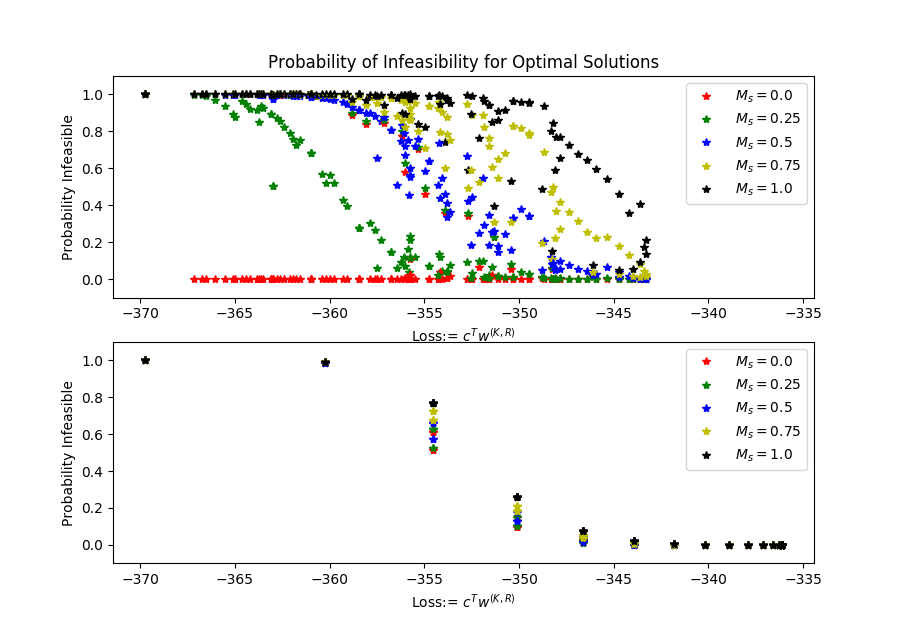} 
 \captionof{figure}{RANDOM: Upper chart plots loss vs. estimated $P(\hat{A}w^{(K,R)} > b)$ for all solutions $w^{(K,R)}$ with $R\neq 0$ for all scenarios $s\in \{1,...,5\}$. Lower chart plots the same for all solutions $w^{(K,R)}$ with $R= 0$. $M_s$ indicates scenario multiplier.}
   \label{fig_A3}

\end{figure}
 In Figure~\ref{fig_1} and \ref{fig_A1}, we see how loss $w^Tc$ changes with growing $k_j$ and
 decreases with shrinking $r_j$.\footnote{Note that the curves in Figure~\ref{fig_1} and \ref{fig_A1} might be expected to be monotone. However, the only reason they are not is because $R$ is set as a \textit{percentage} of $K$ \textit{which is also a percentage}.} Here, we let $w^{(K,R)}$ denote the optimal solution to (\ref{RRLP_2}) when formulated for pair $(K,R)$. This is the expected behavior, since we make sacrifices for robustness $k$ while making gains with optimism $r$. One would expect, though, for increases in $r_j$ to have a large negative impact on each constraints safety margin $(b_j - w^T a_j)$ which provides protection against uncertain increases in the left-hand side of the uncertain constraint. We show, however, that this perceived downside of introducing optimistic $r_j>0$ is not always the case and that for some data sets, the sacrifice in protection for increasing $r_j$ can often be quite small. In Figure~\ref{fig_2} and \ref{fig_A2}, for each of our optimal solutions, we plot the loss relative to the average protection $\frac{1}{\mathcal{J} } \sum_{i=1}^\mathcal{J} (b_j - w^T a_j )$. The arrows along the lines indicate increasing $k_j$. We see in this figure that one can decrease losses by introducing optimistic assumptions with $r_j>0$ without sacrificing much protection on average.

Sometimes the average protection does not tell the full story regarding the probability that your solution will be infeasible for the true coefficient system. In this regard, another perceived downside of introducing optimistic $r_j>0$ lies in the fact that having $r_j>0$ will yield optimal policies $w$ that have no protection against the worst $r_j$ coefficients. Presumably, this lack of protection would work strongly to make $w$ infeasible with high probability. However, in Figure~\ref{fig_3} and \ref{fig_A3} we show that this is not necessarily the case, particularly when your optimistic assumptions are realized, even if only in part. 

Modifying the scheme of \cite{bertsimas2004price}, we plot in Figure~\ref{fig_3} and \ref{fig_A3} the simulated probabilities that our solutions will fail to be feasible once $\hat{A}$ is realized. For these experiments, we need additional notation. First, as before, let $w^{(K,R)}$ denote the optimal solution to (\ref{RRLP_2}) when formulated for pair $(K,R)$. Second, for solution $w^{(K,R)}$ let $\mathcal{I}_{r_j}$ denote the set of indices of the largest $r_j=Rk_j=RK\beta_j$ components of the vector $( |\bar{a}_{1j} w_1^{(K,R)}|,...,|\bar{a}_{nj} w_n^{(K,R)}|)$. Now, recall that when we solve (\ref{RRLP_2}) we are assuming that $\hat{a}_{ij}$ are going to be sampled from the interval $\hat{a}_{ij} \in [a_{ij} -  \bar{a}_{ij}, a_{ij} +  \bar{a}_{ij}]$, \textit{except for coefficients $\hat{a}_{ij}$ with $i \in \mathcal{I}_{r_j}$}. For these coefficients, since $R\neq0$, we have optimistically assumed that they are deterministic with $\hat{a}_{ij} \in [a_{ij} - 0 \cdot\bar{a}_{ij}, a_{ij} + 0\cdot \bar{a}_{ij}]$, i.e. we were able to counteract or mitigate the uncertainty by application of our budget of optimism. To simulate the probability that $w^{(K,R)}$ will be infeasible for the realized coefficient matrix $\hat{A}$, we perform 1000 simulations, counting the number of failures to estimate $P(\hat{a_j}^Tw^{(K,R)} > b_j \text{ for any } j =1,...,\mathcal{J})$. In each simulation, we first sample the value of the random coefficients $\{ \hat{a}_{ij} | i \notin \mathcal{I}_{r_j}, j=1,...,\mathcal{J}\}$ from the worst-case symmetric distribution, which is when $P(\hat{a}_{ij} =a_{ij} -  \bar{a}_{ij}) = P(\hat{a}_{ij} =a_{ij} +  \bar{a}_{ij}) =.5$ and we let the random coefficients $\{ \hat{a}_{ij} | i \in \mathcal{I}_{r_j}, j=1,...,\mathcal{J}\}$ stay fixed at $a_{ij}$. In Figure~\ref{fig_3} and \ref{fig_A3}, the results of these simulations are given by the points with $M_s=0$ in the upper charts.\footnote{The term $M_s$ has a specific meaning which will be explained in the next paragraph.}

However, how will the probability of infeasibility be affected if our optimistic assumptions are not fully realized? In other words, what if the actions taken to mitigate or eliminate the uncertainty are not fully affective, such that the true coefficients $\{ \hat{a}_{ij} | i \in \mathcal{I}_{r_j}, j=1,...,\mathcal{J}\}$ are sampled from $[a_{ij} - M \cdot\bar{a}_{ij}, a_{ij} + M\cdot \bar{a}_{ij}]$ with $0\neq M \leq 1$ instead of staying fixed at $a_{ij}$? To simulate this, we estimate failure probability in 4 additional scenarios $s\in\{1,...,4\}$, each of which specifies the size of the interval that $\hat{a}_{ij}$ will come from via a multiplier $M_s$. Specifically, for each scenario $s$, we let $M_s=.25s$ and perform the same 1000 simulations but where we sample the value of the random coefficients $\{ \hat{a}_{ij} | i \in \mathcal{I}_{r_j}, j=1,...,\mathcal{J}\}$ from the worst-case symmetric distribution with $P(\hat{a}_{ij} =a_{ij} -  M_s\bar{a}_{ij}) = P(\hat{a}_{ij} =a_{ij} +  M_s \bar{a}_{ij}) =.5$. Thus, scenario $s=1,2,3$ indicates that we were able to partially immunize the uncertainty and scenario $s=4$ indicates that our optimistic assumptions were not realized at all and we were not able to immunize any of the uncertainty. In Figure~\ref{fig_3} and \ref{fig_A3}, the results of these simulations are given by the points with $M_s=.25,.5,.75,1$ in the upper charts.

If $R=0$, we have made no optimistic assumptions and simulate with $M_s=1$ and every coefficient coming from the worst case distribution with $P(\hat{a}_{ij} =a_{ij} -  \bar{a}_{ij}) = P(\hat{a}_{ij} =a_{ij} +  \bar{a}_{ij}) =.5$. In Figure~\ref{fig_3} and \ref{fig_A3}, the results of these simulations are given by the points with $M_s=1$ in the lower charts. However, for fair comparison, it is needed to simulate the probability that solutions $w^{(K,0)}$ will be infeasible if, by chance, some optimistic circumstances arose. Therefore, we perform an almost identical procedure for solutions $w^{(K,0)}$. However, let $\mathcal{I}^0_R$ denote the indices of the largest $r_j=Rk_j=RK\beta_j$ components of the vector $( |\bar{a}_{1j} w_1^{(K,0)}|,...,|\bar{a}_{nj} w_n^{(K,0)}|)$. For each solution $w^{(K,0)}$, we then perform the same 1000 iteration simulation for each of 25 scenarios denoted by pairs $(s,R)$ where $s \in \{0,...,3\}$ and $R \in \{0,.05,...,.25\}$. Just as before, we set $M_s$ to encode the varying degrees with which some optimistic circumstance was realized (even though we did not assume that it would happen). Additionally, we encode the occurrences of optimistic circumstances with $R \in \{0,.05,...,.25\}$. In Figure~\ref{fig_3} and \ref{fig_A3}, the results of these simulations are given by the points with $M_s=0,.25,.5,.75$ in the lower charts.

Finally, observing Figure~\ref{fig_3} and \ref{fig_A3}, we can see that when our optimistic assumptions are fully realized (with $M_s=0$), we can substantially decrease losses without sacrificing feasibility by including optimistic uncertainty with $r_j>0$. However, in the case the $M_s \neq0$ where our budget of optimism was not as affective as expected in mitigating uncertainty, we see that there are situations where we still maintain robustness and decrease losses by applying optimism. For example, with $M_s=.25 \text{ or } .5$, we see that we can still maintain feasibility while decreasing losses over the $R=0$ baseline solutions. However, as $M_s$ grows, we see that we indeed begin to become infeasible with high probability. However, this is not too surprising. In this case, we have made optimistic assumptions that were essentially false, i.e. we were not able to mitigate nearly as much uncertainty as we had planned to mitigate. Also, we note that we are assuming the worst case symmetric distribution, so it is not surprising that optimistic solutions are infeasible with high probability if $M_s=1 \text{ or } .75$.

\subsubsection{Convergence of DCA}
\begin{figure}
\centering
\begin{minipage}{.5\textwidth}
  \centering
  \captionsetup{width=.9\linewidth}
  \includegraphics[width=3.24in,height=2.5in]{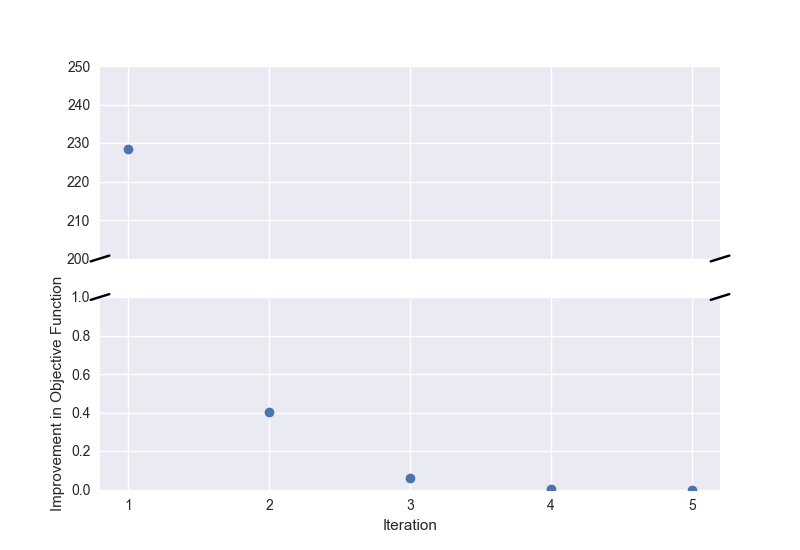} 
 \captionof{figure}{CAPRI}
  \label{fig_iter_capri}

\end{minipage}%
\begin{minipage}{.5\textwidth}

  \centering
  \captionsetup{width=.9\linewidth}
  \includegraphics[width=3.25in,height=2.5in]{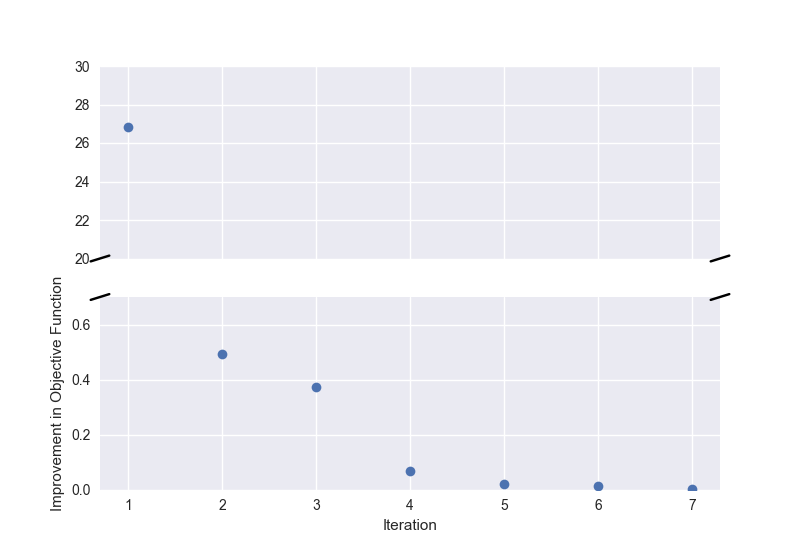} 
  \captionof{figure}{RANDOM}
    \label{fig_iter_random}

\end{minipage}

\end{figure}

Here, we quickly point out one observation regarding the DCA technique utilized to solve the non-convex optimization which was discussed in Section~\ref{sec_algo_OR-LP}. While one of the drawbacks of DCA is the need to solve a convex subproblem at every iteration, we observed in our experiments that very few iterations were needed to reach reasonable solutions, meaning solutions with better objective than the baseline $R=0$ model. In Figure~\ref{fig_iter_capri} and Figure~\ref{fig_iter_random}, we plot the average change in objective value at every iteration, averaged over all runs $(K,R)$, for each data set. For both data sets, we see that the first iteration\footnote{The zeroth iteration involves solving the initial convex problem with $R=0$.}yields large improvement in the objective value, with subsequent iterations yielding much smaller reductions in the overall objective. In general, we see that solutions converge for CAPRI, on average, at around the 4th iteration and for RANDOM around the 5th iteration.

\section{Application to Machine Learning} \label{sec_applyML}
In this section, we discuss application of optimistic uncertainty to various problems in machine learning
and show that the concept of ``optimistic robustness'' is implicitly used in various existing classification and regression models, providing us with a new framework for interpreting and analyzing these approaches. In Section~\ref{sec_sparseML}, we discuss various non-convex regularization strategies, showing that many of them can be shown to be equivalent to optimistic robust optimization strategies. Our primary illustration will utilize binary classification and support vector machines; however, our analysis can also applied to regression problems. Next, in Section~\ref{sec_outliersML} we discuss the use of optimistic robustness for dealing with data corrupted with outliers and noise. Using regression as our illustrative guide, we show that popular techniques for dealing with such problems can be shown to be equivalent to application of optimistic robustness. Overall, we aim to illustrate the breadth of the proposed scheme of optimistic robustness and to motivate future studies which take a more in-depth look into each of these important problem contexts.

\subsection{Non-Convex Regularization is Optimistic Robust} \label{sec_sparseML}
Non-Convex regularization has become a popular technique in statistics and machine learning, often due to its success in inducing sparsity (\cite{fan2011sparse,gong2013general,yin2015minimization,tono2017efficient,gotoh2017dc}) and the ease with which many of these regularizers decompose into DC functions, allowing further use of DC optimization techniques which have proven to be relatively efficient (\cite{tono2017efficient,gotoh2017dc,yin2015minimization}). In this section, we show that non-convex regularizers naturally arise in many contexts when formulating tractable variants of optimistic robust optimization problems. This helps us to intuitively understand why non-convex regularization schemes induce sparsity more efficiently than other methods, such as $L_1$ regularization alone. Specifically, we will focus our attention on the following examples, which have been shown to be representable as DC functions (see \cite{gong2013general}):

\begin{itemize}
\item Approximate $L_0$ norm: $ \|w\|_1 - \rho_k(w)$ (\cite{tono2017efficient,gotoh2017dc})
\item $L_{1-2}$ norm: $ \|w\|_1 - \|w\|_2$ (\cite{yin2015minimization})
\item  Capped $L_1$: $(\theta>0),  \|w\|_1 - \sum_i [|w_i| - \theta]^+ $ (\cite{zhang2010analysis})
\item  MCP: $(\lambda,\theta>0),\lambda \|w\|_1 - \sum_i \begin{cases}\frac{w_i^2}{2\theta} , &: |w_i|\leq \theta \lambda, \\ \lambda |w_i| - \frac{\theta \lambda^2}{2} ,&: |w_i|>\theta \lambda .\end{cases} $ (\cite{zhang2010nearly})
\item SCAD: $(\lambda>0,\theta>2),\lambda \|w\|_1 - \sum_i \begin{cases}0 ,&:|w_i|\leq \lambda, \\ \frac{(|w_i|-\lambda)^2}{2(\theta-1)} ,&: \lambda<|w_i|\leq \theta \lambda, \\  \lambda|w_i| - \frac{(\theta+1)\lambda^2}{2} ,&: |w_i| > \theta \lambda .\end{cases} $ \\(\cite{fan2001variable})
\end{itemize}
We first discuss the Approximate $L_0$ norm and $L_{1-2}$ norm. Specifically, we see that their representation in the optimistic robust framework is straightforward and follows directly from results already discussed. We then discuss separately Capped $L_1$, MCP, and SCAD. We isolate these examples because of their special structure. Specifically, Theorem~\ref{theo_optrobust_nonconv1}, along with most of the results discussed in this paper, rely upon the positive homogeneity of $g$. However, the Capped $L_1$, MCP, and SCAD regularizers are representable as $g^P(w)-g^O(w)$ where $g^O$ is not PH. Is this case, we must take a slightly different approach. Although the approach is different, we are still able to show that these non-convex regularizers intuitively arise from application of an optimistic robust framework.

These non-convex regularizers are typically combined with various loss functions for classification and regression. To illustrate the connection between optimistic robustness and non-convex regularization, we focus on the binary classification problem. Specifically, we show that the popular $\nu$-SVC algorithm from \cite{NECO_Scholkopf+etal_2000}, when formulated with non-convex regularizer, is equivalent to an unregularized optimistic robust optimization problem which simply utilizes the $\nu$-SVC loss function in addition to optimistic robustness w.r.t. the training data. Utilizing this equivalence, we can gain a clear understanding about why such non-convex regularizers induce sparsity in certain optimization problems. We illustrate this graphically in the context of a classification example. 

\subsubsection{Robust representations of non-convex regularizers: PH Case} 
\label{sec_PH_reg}
To show that the Approximate $L_0$ norm and $L_{1-2}$ norm have equivalent representations in an optimistic robust framework, we need only to rely upon a simple transformation. Assuming the same notation as we did in Section~\ref{sec_optrobstopt}, we have that,

$$ \underset{\delta^O \in C(g^O,z^O) }{\text{inf }}   \underset{\delta^P \in C(g^P,z^P)  }{\text{sup }}  w^T ( \delta^P + \delta^O) =  z^Pg^P(w) - z^Og^O(w) \;.
$$
By defining our uncertainty sets in terms of a convex, PH, non-negative support function $g$, we can transform the optimistic robust linear term into the sum of a convex term and a concave term which, considered together, is DC.

Now, consider the case of a binary classification problem, where we have $N$ data observations $(X_i,y_i) \in \mathbb{R}^n \times \{-1,+1\}$ with feature vectors $X_i$ and labels $y_i$, and we need to determine a hyperplane $(w,b)$ that predicts the binary label  $y$ for a new feature vector $X$. The $\nu$-SVC problem from \cite{NECO_Scholkopf+etal_2000} is formulated as
\begin{equation}
\label{nu-SVC}
\begin{aligned}
&\underset{w,b, \epsilon}{\text{min }}  
  & & \|w\|_2^2+\nu \epsilon +\frac{1}{N} \sum_{i=1}^N [ -y_i(w^TX_i +b) -\epsilon]^+ ,\\
\end{aligned}
\end{equation}
where 
$\nu \in (0,1]$ is a hyperparameter chosen apriori.
The term $\nu \epsilon +\frac{1}{N} \sum_{i=1}^N [-y_i(w^TX_i +b) -\epsilon]^+ $
is a type of $\epsilon$-insensitive loss. 
Note that the parameter $\nu \in (0,1]$ equals the fraction of data points laying on or inside the classification margin. 

We first form the unregularized variant, which is simply minimizing the
$\epsilon$-insensitive loss, 
and then, assuming that the feature vectors $X_i$ are uncertain, we arrive at the optimistic robust $\nu$-SVC given by,

\begin{equation}
\label{SVM_risky_robust1}
\begin{aligned}
&\underset{w,b, \epsilon, \delta^O}{\text{min }}   \underset{\delta^P}{\text{max }}  
& &  \nu \epsilon +\frac{1}{N} \sum_{i=1}^N [ -y_i(w^T(X_i+ \delta^O_i + \delta^P_i) +b) -\epsilon]^+ \\
& s.t.  
& &  \delta^O_i \in C(g_i^O,z_i^O) , \delta^P_i \in C(g_i^P,z_i^P), i=1,...,N .\\
&&&
\end{aligned}
\end{equation}

We can then simplify this expression to the following tractable problem (\ref{SVM_risky_robust2}),

\begin{equation}
\label{SVM_risky_robust2}
\begin{aligned}
&\underset{w,b,\epsilon}{\text{min }}  
& &  \nu \epsilon +\frac{1}{N}  \sum_{i=1}^N [ -y_i(w^TX_i+b) -\epsilon   + z_i^P g_i^P(w) - z_i^O g_i^O(w) ]^+ .\\
\end{aligned}
\end{equation}

Let us now assume that each feature vector is subject to the same uncertainty, meaning $g_i^O=g^O, z_i^O=z^O$ and $g_i^P=g^P,z_i^P=z^P$ for all $i=1,...,N$. We can then see that with the correct choice of convex set, non-convex regularizers arise. For example, if we let $z^O=z^P=z$ for some $z \geq 0$ and have uncertainty sets $C(g^O,z)= \{ \delta^O | \|\delta^O\|_2 \leq z\}$, $C(g^P,z)= \{ \delta^P | \|\delta^P\|_\infty \leq z\}$, then (\ref{SVM_risky_robust1}) will reduce to,
\[
 \underset{w,b,\epsilon}{\text{min }} \quad  \nu \epsilon +\frac{1}{N} \sum_{i=1}^N [ -y_i(w^TX_i+b) -\epsilon + z( \|w\|_1 - \|w\|_2  ) ]^+.
\]
By using the variable $\gamma :=\epsilon - z( \|w\|_1 - \|w\|_2  )$  instead of $\epsilon$, we see that this problem becomes the $L_{1-2}$ norm regularized optimization problem, 

\begin{equation}
\label{SVM_risky_robust3}
\begin{aligned}
&\underset{w,b,\gamma}{\text{min }}   
& &  \nu z( \|w\|_1 - \|w\|_2  ) +  \nu \gamma +\frac{1}{N} \sum_{i=1}^N [ -y_i(w^TX_i+b)  -\gamma ]^+ \;.\\
\end{aligned}
\end{equation}
Note that this problem is identical to $\nu$-SVC \eqref{nu-SVC} but with the $L_2$ norm regularizer replaced with the non-convex regularizer $\nu z( \|w\|_1 - \|w\|_2  )$, which is formed by the optimistic robust policy
 of minimizing $\epsilon$-insensitive loss. 
 
Additionally, with appropriate constants and choice of uncertainty set $C(g^O,z)= \{ \delta^O | \rho_k^*(\delta) \leq z \} = \{\delta^O | \| \delta \|_1 \leq zk, \| \delta \|_\infty \leq z\}$, $C(g^P,z)= \{ \delta^P | \|\delta\|_\infty \leq z\}$, we have that (\ref{SVM_risky_robust1}) reduces to the Approximate $L_0$ norm regularized problem,
\begin{equation}
\label{SVM_risky_robust4}
\begin{aligned}
&\underset{w,b,\gamma }{\text{min }} 
& &   \nu z(  \|w\|_1 - \rho_k(w) ) +  \nu \gamma +\frac{1}{N} \sum_{i=1}^N [ -y_i(w^TX_i+b) -\gamma   ]^+ \;.\\
\end{aligned}
\end{equation}
Note that this exact procedure can be performed for regression problems as well.\footnote{Applying this procedure to other loss functions, such as the hinge loss in classification, yield similar, non-convex regularized formulations.} For example, assume we have a regression setting with feature vectors $X_i$ with associated targets $y_i \in \mathbb{R}$ and that we utilize the regression variant of the $\epsilon$-insensitive loss,
\[ \nu \epsilon +\frac{1}{N} \sum_{i=1}^N [ |y_i-(w^TX_i+b)| -\epsilon   ]^+ \;.
\]
Applying the same procedure where we apply the same uncertainty to each $X_i$ and simplify, we get a variant of the regularized $\nu$-SVR algorithm. Specifically, $\nu$-SVR is formulated as
\begin{equation}
\label{nu-SVR}
\begin{aligned}
&\underset{w,b, \epsilon}{\text{min }}  
  & &  \frac{1}{2}\|w\|_2^2 + C(\nu \epsilon +\frac{1}{N} \sum_{i=1}^N [ |y_i-(w^TX_i+b)| -\epsilon   ]^+) \;,\\
\end{aligned}
\end{equation}
where $\nu \in (0,1]$ and $C>0$ are hyperparameters chosen apriori. Applying the proposed uncertainty to $\epsilon$-insensitive loss, we then get the same formulation but with $L_2$ regularizer replaced by non-convex regularizer $\nu z( g^P(w)- g^O(w))$.

\subsubsection{Robust representations of non-convex regularizers: Non-PH Case}
\label{sec_non_PH_reg}
For the Approximate $L_0$ and $L_{1-2}$ norm, we were able to show that this type of regularization arose after simple application of optimistic and pessimistic uncertainty. For Capped $L_1$, MCP, and SCAD, we will show these regularizers naturally arise when we apply optimistic and pessimistic uncertainty \textit{along with an additional penalty applied directly to the optimistic uncertainty.} Specifically, we apply a penalty to the optimistic disturbances $\delta^O$ that discourages particular choices of $\delta^O \in C^O$ where $C^O$ denotes the uncertainty set. While we maintain a binary classification context here, we note that we address the application of this principal in a general context in Appendix~\ref{appendix_B} where we provide a theorem similar in nature to Theorem~\ref{theo_optrobust_nonconv1}.

Consider the following variant of (\ref{SVM_risky_robust1}), where $C_i^O, C_i^P$ are convex sets and $h:\mathbb{R}^n \rightarrow \mathbb{R}$ represents a \textit{penalty} which is incurred for being optimistic:
\begin{equation}
\label{SVM_conjugate}
\begin{aligned}
&\underset{w,b, \epsilon, \delta^O}{\text{min }}   \underset{\delta^P}{\text{max }}  
& &  \nu \epsilon +\frac{1}{N} \sum_{i=1}^N [ -y_i(w^T(X_i+ \delta^O_i + \delta^P_i) +b) + h(\delta_i^O)-\epsilon]^+ \\
& s.t.  
& &  \delta^O_i \in C_i^O , \delta^P_i \in C_i^P, i=1,...,N .\\
&&&
\end{aligned}
\end{equation}

To see how Capped $L_1$, MCP, and SCAD arise out of this formulation, we only need to apply the following identities which are proved in Appendix~\ref{appendix_A}. For brevity, let $M_{\lambda,\theta}(w)$ and $S_{\lambda,\theta}(w)$ denote MCP and SCAD respectively:
\begin{itemize}
\item Capped $L_1$: $ \|w\|_1 - \sum_i [|w_i| - \theta]^+=\underset{\|\delta^P\|_\infty \leq 1}{ \text{sup }}  \underset{\|\delta^O\|_\infty \leq 1}{ \text{inf }}  w^T(\delta^P + \delta^O) + \theta \|\delta^O\|_1\;.$
\item MCP: $M_{\lambda,\theta}(w)=\underset{\|\delta^P\|_\infty \leq \lambda}{ \text{sup }}  \underset{\|\delta^O\|_\infty \leq \lambda}{ \text{inf }}  w^T(\delta^P + \delta^O) + \frac{ \theta}{2} \|\delta^O \|_2^2\;.$
\item SCAD: $S_{\lambda,\theta}(w)=\underset{\|\delta^P\|_\infty \leq \lambda}{ \text{sup }}  \underset{\|\delta^O\|_\infty \leq \lambda}{ \text{inf }}  w^T(\delta^P + \delta^O) + \lambda \|\delta^O\|_1 + \frac{ \theta-1}{2} \|\delta^O \|_2^2\;.$
\end{itemize}

We can then illustrate the use of these identities in (\ref{SVM_conjugate}). For example, for Capped $L_1$, let $h(\delta_i^O)= \theta \|\delta^O\|_1$ and $C_i^O,C_i^P$ each equal the $L_\infty$ norm ball with radius $1$. Then, we see that (\ref{SVM_conjugate}) reduces to,
\begin{equation}
\label{SVM_capped_l1}
\begin{aligned}
&\underset{w,b, \epsilon}{\text{min }}  
& &  \nu \epsilon +\frac{1}{N} \sum_{i=1}^N [ -y_i(w^TX_i+b) + \|w\|_1 - \sum_i [|w_i| - \theta]^+ -\epsilon]^+ \;.
\end{aligned}
\end{equation}
Then, just as we did in Section~\ref{sec_sparseML}, letting $\gamma:= \epsilon - ( \|w\|_1 - \sum_i [|w_i| - \theta]^+)$ we get the Capped $L_1$ regularized $\nu$-SVC:
\begin{equation}
\label{SVM_capped_l1_outside}
\begin{aligned}
&\underset{w,b, \gamma}{\text{min }}  
& &  \nu (\|w\|_1 - \sum_i [|w_i| - \theta]^+) +\nu \gamma +\frac{1}{N} \sum_{i=1}^N [ -y_i(w^TX_i+b) - \gamma]^+ \\
\end{aligned}
\end{equation}
Using this procedure, it is easy to see that similar results hold for MCP and SCAD, with these non-convex penalty terms arising naturally out of optimistic robust optimization problems. 

One will notice that the optimistic robust representations of Capped $L_1$, MCP, and SCAD are all essentially identical except for their choice of \textit{optimistic disturbance penalty} $h$, i.e. the way in which they penalize optimism. Specifically, we see that Capped $L_1$ utilizes an $L_1$ penalty with weight $\theta$, MCP utilizes a squared $L_2$ penalty with weight $\theta$, and SCAD utilizes an elastic-net penalty\footnote{See \cite{zou2005regularization}.} with weights $\lambda,\theta$.
\subsubsection{Visualizing Non-Convex Regularization}
Now that we have seen that these non-convex regularizers can be interpreted as the combination of optimistic and pessimistic uncertainty, we can use this interpretation to help understand why these non-convex regularizers induce sparsity more strongly than, say, the lone $L_1$ norm. We utilize a visualization to help foster intuition.
 
\begin{figure}
  \centering
  \captionsetup{width=.9\linewidth}
  \includegraphics[width=5in,height=2.75in]{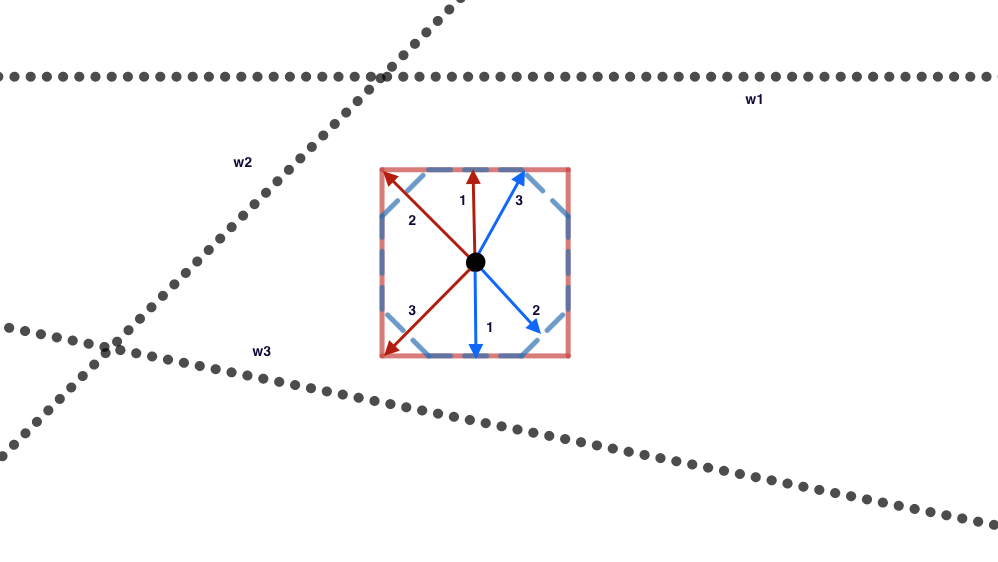} 
 \captionof{figure}{Illustration of classification with regularizer $ \|w\|_1 - \rho_k(w)$. Dotted lines represent three different classification hyperplanes $w_1,w_2,w_3$ that all correctly classify the illustrated data point $X$ (the large black dot). Boundary of uncertainty sets $\{\delta^O | \rho_k^*(\delta^O) \leq 1\}, \{\delta^P | \| \delta^P\|_\infty \leq 1\}$ represented by blue and red polygons. Given selection of $w_i$ as the classification hyperplane, $i^{th}$ blue and red vectors represent respectively the resulting optimistic $\delta^O$ and pessimistic $\delta^P$ disturbances for SVC objective. }
  \label{fig_4}
\end{figure}

In Figure~\ref{fig_4}, we visualize in two dimensions the uncertainty set that would be around a single feature vector $X$ in the classification problem (\ref{SVM_risky_robust4}) and the effect of the competing forces, optimism and pessimism. In this figure, the dotted lines represent three different classification hyperplanes $w_1,w_2,w_3$ that all correctly classify the illustrated data point $X$. Here, $w_1$ is our sparse hyperplane (having a single zero component) while $w_2,w_3$ are non-sparse. The boundary of uncertainty sets $C(g^O,1)= \{ \delta^O | \rho_k^*(\delta^O) \leq 1 \} = \{\delta^O | \| \delta^O \|_1 \leq k, \| \delta^O \|_\infty \leq 1\}$ and $C(g^P,1)= \{ \delta^P | \|\delta^P\|_\infty \leq 1\}$ are represented by blue and red polygons respectively. 

We begin by first assuming that $w_1$ is going to be our selected hyperplane and asking: Given selection of $w_1$ as the regression hyperplane, what will $\delta^O$ and $\delta^P$ equal? We visualize one of the possible solutions with the blue and red arrows labeled with the number 1. Specifically, we know that by construction of the classification problem, the pessimistic $\delta^P$ will attempt to move $X$ as close to $w_1$ as possible. Conversely, the optimistic $\delta^O$ will move $X$ as far away from $w_1$ as possible.  We then see that the resulting arrows cancel out, moving opposite each other in the direction of the normal vector $w_1$ and its negative. 

What happens, then, if we were to select either $w_2$ or $w_3$ as our classification hyperplane? For $w_i$, we visualize this with the $i^{th}$ blue and red vectors representing the resulting optimistic $\delta^O$ and pessimistic $\delta^P$ disturbances for SVC objective. Notice here that the construction of the uncertainty sets has a noticeable affect. In the case of the sparse $w_1$, because the hyperplane was parallel to the face of the $L_\infty$ uncertainty set, we had $\delta^O$ and $\delta^P$ cancel each other out. For $w_2$ and $w_3$, though, this is not the case. In both scenarios, we find that the magnitude of $\delta^P$ is larger than the magnitude of $\delta^O$, and thus the objective will be hurt by the pessimistic uncertainty which \textit{wins} against the optimism. We can then see that this uncertainty set, by its construction, incentivizes sparse hyperplanes. If the hyperplane has a non-sparse dimension, then the pessimism will dominate the optimism and hurt the objective function.

Finally, although this illustration only addressed a case from Section~\ref{sec_PH_reg}, the same intuition can be applied to the cases discussed in Section~\ref{sec_non_PH_reg}. Specific penalty terms $h(\delta^O)$ can be used to incentivize the selection of sparse hyperplanes. By enforcing a larger penalty in the corner of the $L_\infty$ box and smaller penalty at the midpoint of a side (e.g. with $h(\delta^O)=\|\delta^O\|_1$ as is done with Capped $L_1$), sparse hyperplanes will be incentivized because the optimistic disturbance will have more power (incur less of a penalty) if the hyperplane is indeed sparse and thus the optimistic disturbance will be like the $1^{st}$ blue vector in Figure~\ref{fig_4}. 

\subsection{Fighting Outliers and Noisy Data with Optimism} \label{sec_outliersML}
Although an optimistic approach to uncertainty may sound unusual, it can be quite practical for real-life situations. In this section, we show that this approach is connected to existing regression and classification approaches for dealing with data corrupted with noise or outliers. 


\subsubsection{Outliers}\label{sec_robust_outliersML}

A situation in which an optimistic view of uncertainty can be useful is in the case of training a learner on a data set with outliers. The presence of a few outliers can have a dramatic effect on learning algorithms and there has been a great deal of research done to address these difficulties across multiple learning tasks. For the case of regression, see literature on classical robust statistics\footnote{In statistics, robustness usually refers to insensitivity to outliers, different from the concept of robust optimization assuming the data input to be uncertain and to belong to some fixed uncertainty set. There have been a large number of works on classical robust statistics, which develop estimation methods that are robust to outliers.} such as \cite{rousseeuw1987LTS} and the relatively recent work of \cite{yu2010NIPS},
and for the case of classification see e.g. \cite{xu2006robust,takeda2014extended,tsyurmasto2014value,fujiwara2017dc}.

We can approach this problem by taking an optimistic viewpoint of the data. After formulating an optimistic robust model to address this problem, we will see that it is quite similar to existing methods in robust statistics for dealing with outliers. We maintain a regression context, however, this technique has obvious application to classification as well. 

By combining Proposition~8 and Theorem~10 of \cite{JBF:Rockafellar+Uryasev:2002}, we see that for any fixed $(w,b)$,
the optimal value of 
\[
\underset{\epsilon}{\text{min }}  \epsilon +\frac{1}{\nu N} \sum_{i=1}^N [ |y_i-(w^T X_i +b)| -\epsilon]^+ 
\]
is equal to
\begin{align*}
\rho_{\nu N}(z)&=\frac{1}{k} (\sum_{i=1}^{\lfloor k \rfloor} |z^{(i)} | + (k - \lfloor k \rfloor)|z^{(\lceil k \rceil)}|) \\
&= \max_{\rho_{\nu N}^*(\delta) \leq 1} z^T\delta \;,  
\end{align*}
where $k=\nu N$,
the vector $z$ is defined by components\footnote{$z$ depends on the variable $(w,b)$ but we write $z$ for simplicity of notation.}  $z_i:=y_i-(w^T X_i +b)$, $\forall i$, and the
ordered components (w.r.t. absolute value) of $z$ are denoted as $|z^{(1)}| \geq \dots \geq |z^{(N)}|$.
Note that the above problem corresponds to minimizing the $\epsilon$-insensitive loss function term in $\nu$-SVR  \eqref{nu-SVR}.
Therefore,  $\nu$-SVR \eqref{nu-SVR} can be reformulated as
\begin{equation}
\label{nu-SVR_dual}
\begin{aligned}
&\underset{w,b}{\text{min }}  \underset{\delta}{\text{max }}  
  & & \frac{1}{2}\|w\|_2^2+ \frac{C}{N} z^T\delta \\
  & s.t.  
  & & \|  \delta \|_1 \leq \nu N, ~ \|  \delta \|_\infty \leq 1\;,\\
\end{aligned}
\end{equation}
whose feasible set can also be written as $\rho_{\nu N}^*(\delta) \leq 1$. Equivalently, this can be formulated as,
\begin{equation*}
\begin{aligned}
&\underset{w,b}{\text{min }}  
  & & \frac{1}{2}\|w\|^2+ \frac{C}{N} \rho_{\nu N}(z). \\
\end{aligned}
\end{equation*}

\begin{figure}
\centering
  \centering
  \captionsetup{width=.9\linewidth}
  \includegraphics[width=3.24in,height=2.5in]{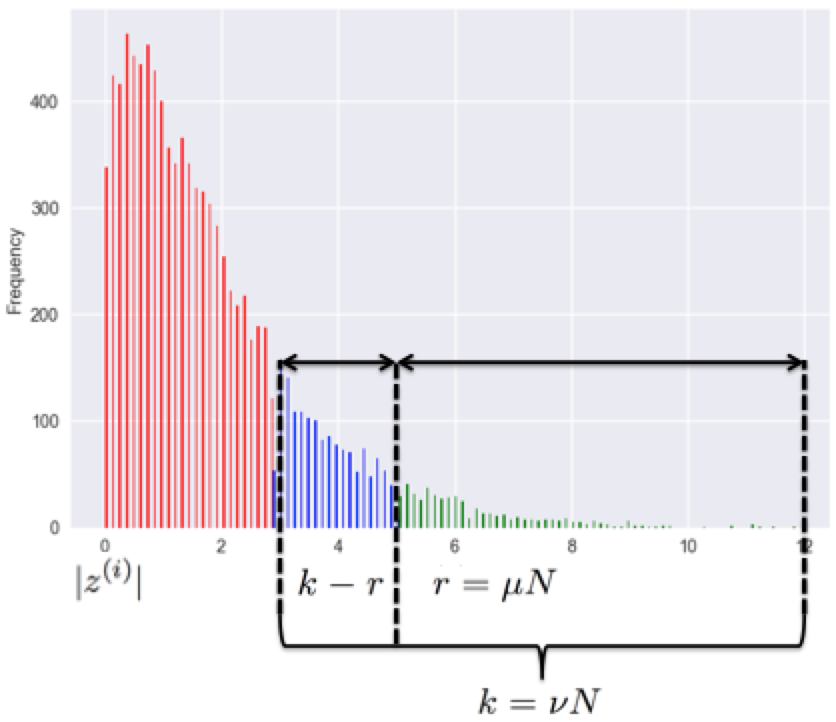} 
  \captionof{figure}{ Distribution of $|z^{(i)}|$, $\forall i$. The optimistic robust model minimizes the average of the $|z^{(i)}|$'s in the blue area, ignoring the $r$ largest errors in green (presumably caused by the outliers).}
 \label{fig_hist}
\end{figure}

Here, we apply the optimistic strategy to $\nu$-SVR \eqref{nu-SVR_dual} with additional parameter $0< \mu < \nu$ by following a strategy similar to the one applied in OR-LP \eqref{RRLP_1}, yielding
\begin{equation*}
\begin{aligned}
&\underset{w,b, \delta^O}{\text{min }}  \underset{\delta^P}{\text{max }}  
  & & \frac{1}{2}\|w\|^2+   \frac{C}{N}   z^T ( \delta^O+\delta^P)\\ 
  & s.t.  
  & & \rho_{\nu N}^*(\delta^P) \leq 1,~  \rho_{\mu N}^*(\delta^O) \leq 1 \;,
\end{aligned}
\end{equation*}
which is equivalent to minimizing
\begin{equation}
    \label{regression_OR-LP}
\begin{aligned}
\frac{1}{2}\|w\|^2+   \frac{C}{N} (\rho_{\nu N}(z) -\rho_{\mu N}(z) ).
\end{aligned}
\end{equation}
By expressing $\rho_{\nu N}(z)$ and $\rho_{\mu N}(z)$ by 
 optimization problems \eqref{cvar_opt}, we have a tractable problem for  \eqref{regression_OR-LP}, though it is a non-convex optimization problem. This problem can be posed equivalently as the more interpretable problem,
\begin{equation*}
\begin{aligned}
&\underset{w,b}{\text{min }}  
  & & \frac{1}{2}\|w\|^2+   \frac{C}{N} 
\left(  \sum_{i=\lfloor r \rfloor+1}^{\lfloor k \rfloor} |z^{(i)} | + (k - \lfloor k \rfloor)|z^{(\lceil k \rceil)}|   - (r - \lfloor r \rfloor)|z^{(\lceil r \rceil)}| \right)  \;,\\
\end{aligned}
\end{equation*}
where $k=\nu N$ and $r=\mu N$. Figure~\ref{fig_hist} explains the second term of this objective function. Note that the largest $\lfloor \mu N \rfloor$ errors, that is, the $z^{(i)}$ with $i=1,\ldots,\lfloor \mu N \rfloor$, are considered outliers and ignored in the optimistic robust formulation.

Various non-convex regression models have been studied with the goal of ensuring robustness to outliers. They are often called ``robust'' regression models with the Least Trimmed Squares (LTS) of \cite{rousseeuw1987LTS} being a good representative of these models where LTS minimizes $\sum_{i=\lfloor r \rfloor+1}^N (z^{(i)})^2$. When we define $z_i=(y_i-(w^T X_i +b))^2$ and set $\nu=1$, the resulting optimistic robust model \eqref{regression_OR-LP} equals LTS. Recent robust statistics models (see e.g. \cite{xu2006robust,yu2010NIPS}) use truncated (or trimmed, clipped) loss functions that are defined as $\min\{\ell(X_i,y_i;w,b),\tau\}$ with a threshold $\tau (>0)$ for some convex loss function $\ell(\cdot)$. The truncated loss function can bound the influence of any outliers on the final result. Minimizing the truncated loss function defined with loss $\ell(X_i,y_i;w,b)$ can find the same optimal solution $(w,b)$ to  OR-LP \eqref{regression_OR-LP} with $z_i=\ell(X_i,y_i;w,b)$ if the value $\tau$ of the truncated model and the ratio $\mu$ of OR-LP are set appropriately.


\subsubsection{Noisy Data} \label{sec_noisyML}
In this subsection, we again consider the standard binary classification problem. A standard regularized classifier that is counterpart to the $\nu$-SVC is the well known $C$-SVM of \cite{cortes1995support} which yields classifier $(w, b)$ by solving the following convex optimization problem:
\begin{equation}
\label{SVM_hinge}
\begin{aligned}
&\underset{w,b}{\text{min }}  
& & C \sum_{i=1}^N [ -y_i(w^TX_i +b) +1]^+ \; + \frac{1}{2}\|w\|^2.\\
\end{aligned}
\end{equation}
For binary classification problems, there are existing works which take two different strategies to deal with uncertainty of given data samples. One is an optimistic strategy based on the idea that even if there is a large margin separator for the original uncorrupted inputs, the observed noisy data may become non-separable. \cite{zhang2005support} took the optimistic strategy that tries to recover the original classifier from the corrupted training data. On the other hand, \cite{xu2009robustness,trafalis2007robust} took another strategy that is pessimistic with regard to the data uncertainty. They assume that data uncertainties and perturbations are unknown but the perturbed data are supposed to be living in some certain sets. Their aim is to obtain classifiers which have good generalization properties and are insensitive to bounded perturbations of the data. 

Consider the optimistic situation where one must train a learning algorithm on data that has been contaminated with noise, knowing that at test-time, the input data will not be contaminated with such noise. In this case, it can be beneficial to train your learning algorithm on an optimistic version of the training data, which can be accomplished via optimistic uncertainty.
In fact, it is easy to see that the technique utilized by \cite{zhang2005support}, which yields the Total Support Vector Classifier (TSVC), is equivalent to a special case of optimistic robust optimization. Specifically, the TSVC is given by: 
\begin{equation}
\label{TSVM}
\begin{aligned}
  &\underset{w,b,\delta^O}{\text{min }}
  & & C \sum_{i=1}^N [ -y_i(w^T(X_i+ \delta_i^O) +b) +1]^+ \; + \frac{1}{2}\|w\|^2\\
& s.t.  
& & \| \delta_i^O\|_2 \leq z_i^O, i=1,...,N \;,
\end{aligned}
\end{equation}
where a simple bounded uncertainty $\| \delta_i^O\|_2 \leq z_i^O$ is assumed.
The problem can be transformed into a non-convex optimization minimizing the non-convex function:
\[
C \sum_{i=1}^N [ -y_i(w^T X_i +b) -  z_i^O \|w\| +1]^+ \; + \frac{1}{2}\|w\|^2
\]
with respect to $(w,b)$ and is solved by \cite{zhang2005support} by iteratively solving a series of $C$-SVMs.

This is in contrast to the pessimistic approach of \cite{trafalis2007robust}, which minimizes the worst-case empirical error under the given model of uncertainty with formulation given by,
\begin{equation}
\label{RobSVM}
\begin{aligned}
  &\underset{w,b}{\text{min }} \underset{\delta^P}{\text{max }}  
  & & C \sum_{i=1}^N [ -y_i(w^T(X_i+ \delta_i^P) +b) +1]^+ \; + \frac{1}{2}\|w\|^2\\
& s.t.  
& & \| \delta_i^P\|_2 \leq z_i^P, i=1,...,N \;.
\end{aligned}
\end{equation}
The problem can be transformed into a convex optimization, specifically the second-order cone programming (SOCP) problem of minimizing the convex function:
\[
C \sum_{i=1}^N [ -y_i(w^T X_i +b) +  z_i^P \|w\| +1]^+ \; + \frac{1}{2}\|w\|^2
\]
with respect to $(w,b)$. This problem can be efficiently solved by an interior-point method.

Moreover, \cite{xu2009robustness} has shown that the regularization term $\|w\|^2$ naturally comes from non-regularized robust $C$-SVM by assuming some specific type of uncertainty sets for $\delta_i^P$, just as the non-convex regularizer $\|w\|_1 - \|w\|_2 $ naturally appears in \eqref{SVM_risky_robust3}. Also, \cite{NortonSVM} showed that the non-convex regularization of the T-SVC naturally arises via optimistic robustness applied to the non-regularized C-SVM. The link between regularization and robustness in classification suggests that norm-based regularization essentially builds in a robustness to sample noise whose probability level sets are symmetric.

Thus, we see that application of optimistic uncertainty is a natural consideration in the context of making predictions with noisy training data. Indeed, we see that we can make a direct connection between the optimistic strategy and existing methods (the TSVC). Taking into consideration the existence of the two approaches, both pessimistic and optimistic, it would then seem natural to consider the mixed approach like that considered in Section~\ref{sec_sparseML}. Although we leave a full exploration to future work, it would seem intuitive that the mixed strategy, via proper construction of the optimistic and pessimistic uncertainty set, could mitigate the extreme decisions made by using a single strategy alone. Clearly, by applying the specific mixed strategy of Section~\ref{sec_sparseML}, we would achieve sparsity inducing non-convex regularization. However, it is not clear which uncertainty sets should be chosen to optimally combat noisy data. \\

\noindent \textbf{Remark:} One must be careful when applying both robustness and regularization simultaneously. In some cases, though it may not be obvious, the approaches are redundant. For example, it can be shown via analysis of KKT conditions that
\[
\min_{w,b} \quad C \sum_{i=1}^N [ -y_i(w^T X_i +b) +  z^P \|w\| +1]^+ \; + \frac{1}{2}\|w\|^2
\]
is equivalent to
\[
\min_{w,b} \quad \hat{C} \sum_{i=1}^N [ -y_i(w^T X_i +b)  +1]^+ \; + \frac{1}{2}\|w\|^2
\]
for the proper choice of $\hat{C}$ if the same norm is used inside and outside of the hinge loss.\footnote{See, for example, Theorem 1, Theorem 2, and Corollary 1 in \cite{NortonSVM}.} 
\section{Conclusion}
Robust Optimization traditionally assumes a pessimistic attitude towards uncertainty, focusing purely on the worst-case outcome. We have attempted to provide convincing argument that it can often be beneficial to consider an optimistic attitude towards uncertainty. This optimistic, best-case viewpoint can provide a mechanism for reducing the conservatism of traditional, pessimistic formulations in intuitive ways that can easily be connected with realistic motivation. For example, if the decision maker has at his disposal a set of resources which can be used to combat, or mitigate unexpected realizations of model parameters, then optimistic uncertainty can be utilized to represent these circumstances and suggest optimal ways of allocating these resources (the budget of optimism). Additionally, when mixed-in with pessimistic assumptions, we are able to obtain profitable solutions that are still robust to sets of circumstances that go beyond our budget of optimism.

We have also attempted to provide justification for the optimistic view of uncertainty by connecting this framework with recent developments in machine learning and statistics. Focusing on regularization, we were able to show that non-convex, sparsity inducing regularizers can sometimes be interpreted as a simple application of optimistic robust optimization. This new result allows us to provide a new interpretation for these non-convex regularizers, specifically providing a geometric explanation for why these regularization strategies indeed induce sparsity. Moving beyond regularization, we were also able to show that popular methods for dealing with outliers and noisy training data can be viewed as special application of the general optimistic robust framework.

Overall, we see that the optimistic view of uncertainty yields useful strategies for solving a variety of problems with formulations that can often be solved via simple optimization strategies (i.e. DCA). Additionally, with strong connections to proven methods in machine learning and statistics, the optimistic view of uncertainty certainly merits attention as a general tool to be applied in tandem with traditional robust optimization. 

\section*{Acknowledgements}
This work was partially supported by the International Internship Support Program of the Institute of Statistical Mathematics, ROIS, and Grant-in-Aid for Scientific Research (C), 15K00031, and USA Air Force Office of Scientific Research grant: ``Design and Redesign of Engineering Systems", FA9550-12-1-0427, and  ``New Developments in Uncertainty: Linking Risk Management, Reliability, Statistics and Stochastic Optimization", FA9550-11-1-0258.
\appendix
\section{Non-convex regularizer identites}
\label{appendix_A}
Here, we show that Capped $L_1$, MCP, and SCAD can be represented in a way amenable to robust reformulations. For all cases, we rely heavily on the fact that the convex conjugate of a convex function $g$ is given by $g^*(w)=\sup_\delta w^T\delta - g(\delta)$ and, furthermore, if $g$ is closed-convex we have $g(w)=\sup_\delta w^T\delta - g^*(\delta)$.
\subsection{Capped $L_1$}
Consider the Capped $L_1$ penalty from \cite{zhang2010analysis}, which is given in DC form by \cite{gong2013general} as,
$$g^P(w)-g^O(w)=\|w\|_1 - \sum_i [|w_i| - \theta]^+ \;,$$
with parameter $\theta >0$. Let $f(x)=[|x|-\theta]^+$. It can be shown that the conjugate of $f$ is given by,
\begin{align*}
f^{*}(d)  &= \sup_x xd - f(x) \\
&=\begin{cases}\infty ,&: |d| >1, \\ |d| \theta ,&: |d| \leq 1.  \end{cases} 
\end{align*}
Since $g^O(w)=\sum_i f(w_i)$ is a separable sum, we know that its conjugate is $g^{*O}(\delta)= \sum_i f^*(\delta_i)$. Using this result, we have that,
\begin{align*}
 g^{O}(w) = \sup_\delta w^T\delta -g^{*O}(\delta) &= \sup_\delta w^T\delta -\sum_i f^*(\delta_i) \\
 &= \sup_\delta w^T\delta -\sum_i \begin{cases}\infty ,&: |\delta_i| >1, \\ |\delta_i|\theta ,&: |\delta_i|\leq 1.  \end{cases} \\
 &= \sup_{\|\delta\|_\infty \leq 1} w^T\delta -\theta \|\delta\|_1 \\
  &= - \inf_{\|\delta\|_\infty \leq 1} \theta \|\delta\|_1  - w^T\delta \\
    &= - \inf_{\|\delta\|_\infty \leq 1} \theta \|\delta\|_1  + w^T\delta \;.
 \end{align*}
Since the first term $\|w\|_1=\sup_{\|\delta\|_\infty \leq 1} w^T\delta$, we can then rewrite the Capped $L_1$ penalty as,
\begin{align*}
g^P(w)-g^O(w) &=\|w\|_1 - \sum_i [|w_i| - \theta]^+ \\
&= \|w\|_1 -  \sup_\delta w^T\delta -g^{*O}(\delta) \\
&= \|w\|_1 + \inf_{\|\delta\|_\infty \leq 1} \theta \|\delta\|_1  + w^T\delta \\
&= \lp( \sup_{\|\delta^P\|_\infty \leq 1} w^T\delta^P \rp)+ \lp( \inf_{\|\delta^O\|_\infty \leq 1} w^T\delta^O + \theta \|\delta^O\|_1\rp) \\
&=\sup_{\|\delta^P\|_\infty \leq 1}  \inf_{\|\delta^O\|_\infty \leq 1} w^T(\delta^P + \delta^O) + \theta \|\delta^O\|_1 \;.
\end{align*}

\subsection{MCP}
We use a proof structure identical to the one followed for Capped $L_1$. Consider the MCP penalty of \cite{zhang2010nearly}, which is given in DC form by \cite{gong2013general} as,
$$g^P(w)-g^O(w)= \lambda \|w\|_1 - \sum_i \begin{cases}\frac{w_i^2}{2\theta} ,&: |w_i|\leq \theta \lambda, \\ \lambda |w_i| - \frac{\theta \lambda^2}{2} ,&: |w_i|>\theta \lambda. \end{cases} $$
with parameters $\lambda>0,\theta>0$.
Let $f(x)=\begin{cases}\frac{x^2}{2\theta} ,&: |x|\leq \theta \lambda, \\ \lambda |x| - \frac{\theta \lambda^2}{2} ,&: |x|>\theta \lambda. \end{cases}$ It can be shown that the conjugate of $f$ is given by,
\begin{align*}
f^{*}(d)  &= \sup_x xd - f(x) \\
&=\begin{cases}\infty ,&: |d| >\lambda, \\ \frac{d^2 \theta}{2} ,&: |d| \leq \lambda.  \end{cases} 
\end{align*}
Since $g^O(w)=\sum_i f(w_i)$ is a separable sum, we know that its conjugate is $g^{*O}(\delta)= \sum_i f^*(\delta_i)$. Using this result, we have that,
\begin{align*}
 g^{O}(w) = \sup_\delta w^T\delta -g^{*O}(\delta) &= \sup_\delta w^T\delta -\sum_i f^*(\delta_i) \\
 &= \sup_\delta w^T\delta -\sum_i \begin{cases}\infty ,&: |\delta_i| >\lambda, \\ \frac{\delta_i^2 \theta}{2} ,&: |\delta_i| \leq \lambda.  \end{cases} \\
 &= \sup_{\|\delta\|_\infty \leq \lambda} w^T\delta - \sum_i \frac{\delta_i^2 \theta}{2}  \\
  &= \sup_{\|\delta\|_\infty \leq \lambda} w^T\delta - \frac{ \theta}{2} \|\delta \|_2^2   \\
  &= - \inf_{\|\delta\|_\infty \leq \lambda} \frac{ \theta}{2} \|\delta \|_2^2  - w^T\delta \\
    &= - \inf_{\|\delta\|_\infty \leq \lambda} \frac{ \theta}{2} \|\delta \|_2^2  + w^T\delta \;.
 \end{align*}
Since the first term $\lambda \|w\|_1=\sup_{\|\delta\|_\infty \leq \lambda} w^T\delta$, we can then rewrite the MCP penalty as,
\begin{align*}
g^P(w)-g^O(w) &= \lambda \|w\|_1 -  \sup_\delta w^T\delta -g^{*O}(\delta) \\
&=\lambda \|w\|_1 + \inf_{\|\delta\|_\infty \leq \lambda} \frac{ \theta}{2} \|\delta \|_2^2  + w^T\delta \\
&= \lp( \sup_{\|\delta^P\|_\infty \leq \lambda} w^T\delta^P \rp)+ \lp( \inf_{\|\delta^O\|_\infty \leq \lambda} w^T\delta^O + \frac{ \theta}{2} \|\delta^O \|_2^2\rp) \\
&=\sup_{\|\delta^P\|_\infty \leq \lambda}  \inf_{\|\delta^O\|_\infty \leq \lambda} w^T(\delta^P + \delta^O) + \frac{ \theta}{2} \|\delta^O \|_2^2 \;.
\end{align*}

\subsection{SCAD}
We use a proof structure identical to the one followed for Capped $L_1$ and MCP. Consider the SCAD penalty of (\cite{fan2001variable}), which is given in DC form by \cite{gong2013general} as,
$$g^P(w)-g^O(w)= \lambda \|w\|_1 -\sum_i \begin{cases}0 ,&: |w_i|\leq \lambda \\ \frac{(|w_i|-\lambda)^2}{2(\theta-1)} ,&: \lambda<|w_i|\leq \theta \lambda, \\  \lambda|w_i| - \frac{(\theta+1)\lambda^2}{2} ,&: |w_i| > \theta \lambda. \end{cases} \;.$$
with parameters $\lambda>0,\theta>2$.
Let $f(x)=\begin{cases}0 ,&: |x|\leq \lambda, \\ \frac{(|x|-\lambda)^2}{2(\theta-1)} ,&: \lambda<|x|\leq \theta \lambda, \\  \lambda|x| - \frac{(\theta+1)\lambda^2}{2} ,&: |x| > \theta \lambda. \end{cases}$. It can be shown that the conjugate of $f$ is given by,
\begin{align*}
f^{*}(d)  &= \sup_x xd - f(x) \\
&=\begin{cases}\infty ,&: |d| >\lambda, \\ \frac{d^2 (\theta-1)}{2} + \lambda|d|,&: |d| \leq \lambda.  \end{cases} 
\end{align*}
Since $g^O(w)=\sum_i f(w_i)$ is a separable sum, we know that its conjugate is $g^{*O}(\delta)= \sum_i f^*(\delta_i)$. Using this result, we have that,
\begin{align*}
 g^{O}(w) = \sup_\delta w^T\delta -g^{*O}(\delta) &= \sup_\delta w^T\delta -\sum_i f^*(\delta_i) \\
 &= \sup_\delta w^T\delta -\sum_i \begin{cases}\infty ,&: |\delta_i| >\lambda, \\ \frac{\delta_i^2 (\theta-1)}{2} + \lambda|\delta_i|,&: |\delta_i| \leq \lambda.  \end{cases} \\
 &= \sup_{\|\delta\|_\infty \leq \lambda} w^T\delta - \sum_i \lp( \frac{\delta_i^2 (\theta-1)}{2}  + \lambda |\delta_i| \rp)  \\
  &= \sup_{\|\delta\|_\infty \leq \lambda} w^T\delta -\lambda \|\delta\|_1 - \frac{ \theta-1}{2} \|\delta \|_2^2   \\
  &= - \inf_{\|\delta\|_\infty \leq \lambda} \lambda \|\delta\|_1 + \frac{ \theta-1}{2} \|\delta \|_2^2  - w^T\delta \\
    &= - \inf_{\|\delta\|_\infty \leq \lambda} \lambda \|\delta\|_1 + \frac{ \theta-1}{2} \|\delta \|_2^2  + w^T\delta \;.
 \end{align*}
Since the first term $\lambda \|w\|_1=\sup_{\|\delta\|_\infty \leq \lambda} w^T\delta$, we can then rewrite the SCAD penalty as,
\begin{align*}
g^P(w)-g^O(w) &= \lambda \|w\|_1 -  \sup_\delta w^T\delta -g^{*O}(\delta) \\
&=\lambda \|w\|_1 + \inf_{\|\delta\|_\infty \leq \lambda}\lambda \|\delta\|_1 + \frac{ \theta-1}{2} \|\delta \|_2^2   + w^T\delta \\
&= \lp( \sup_{\|\delta^P\|_\infty \leq \lambda} w^T\delta^P \rp)+ \lp( \inf_{\|\delta^O\|_\infty \leq \lambda} w^T\delta^O + \lambda \|\delta^O\|_1 + \frac{ \theta-1}{2} \|\delta^O \|_2^2 \rp) \\
&=\sup_{\|\delta^P\|_\infty \leq \lambda}  \inf_{\|\delta^O\|_\infty \leq \lambda} w^T(\delta^P + \delta^O) + \lambda \|\delta^O\|_1 + \frac{ \theta-1}{2} \|\delta^O \|_2^2  \;.
\end{align*}

\section{Penalizing Disturbances}
\label{appendix_B}
Encapsulating the general flavor of the approach wherein optimistic and pessimistic uncertainty are included \textit{along with an additional penalty applied to optimistic disturbances}, we have the following theorem which is similar in nature to Theorem~\ref{theo_optrobust_nonconv1}. For the following theorem, recall that the convex conjugate of a convex function $g$ is given by $g^*(\delta)=\sup_x x^T\delta - g(x)$. If $g$ is closed-convex, we also have the relation $g(w)=\sup_\delta w^T\delta - g^*(\delta)$. 
\begin{thm}\label{theo_optrobust_nonconv2}
If $g_i^O, g_i^P, i=1,...,\ell$ are closed convex functions, then \eqref{LP_risky_robust2} can be reformulated as,
\begin{equation}
\label{LP_risky_robust_conjugate}
\begin{aligned}
&\underset{w,\delta^O}{\text{min }}   \underset{\delta^P}{\text{max }}  
& & w^T(X_0+ z_0^O\delta^O_0 + z_0^P\delta^P_0) +z_0^Og_0^{*O}(\delta^O_0) - z_0^P g_0^{*P}(\delta^P_0) \\
& s.t.  
& & w^T(X_i +  z_i^O\delta^O_i +  z_i^P\delta^P_i) +z_i^Og_i^{*O}(\delta^O_i) - z_i^P g_i^{*P}(\delta^P_i) \leq 0 , i=1,...,\ell \\
\end{aligned}
\end{equation}
where $g^*$ denotes the convex conjugate of $g$.
\end{thm}
The result follows by simply letting $g(w)=\sup_\delta w^T\delta - g^*(\delta)$ in \eqref{LP_risky_robust2} and making the appropriate manipulations. Note that the emphasis of this result is that it relaxes the assumption of positive homogeneity made in Theorem~\ref{theo_optrobust_nonconv1}. Furthermore, this result relates to the discussion in Section~\ref{sec_non_PH_reg} by the fact that Capped $L_1$, MCP, and SCAD yield conjugates in the form of $\sup_\delta w^T\delta -\sum_i \begin{cases}\infty &,: |\delta_i| >\lambda, \\ h_i(\delta_i) ,&: |\delta_i| \leq \lambda.  \end{cases}$, which can be simplified to $\sup_{\|\delta\|_\infty \leq \lambda} w^T\delta -\sum_i h_i(\delta_i)$.

\bibliographystyle{plainnat} 
\bibliography{RiskyRobustBibFile}

\end{document}